\documentclass{tlp}
\usepackage{latexsym}
\usepackage{color}
\usepackage{oldlfont,rawfonts}
\usepackage{graphicx}
\usepackage{verbatim}

\def \casi{\left \{\begin{array}{ll}}

\def \bottom{\perp}

\newcommand {\bes} {\begin{description}}
\newcommand{\ens} {\end{description}}

\newcommand {\la} {\langle}
\newcommand {\ra} {\rangle}

\newcommand {\orr} {\vee}

\newcommand {\bit} {\begin{itemize}}
\newcommand {\enit} {\end{itemize}}

\newcommand {\beq} {\begin{quote}}
\newcommand {\enq} {\end{quote}}

\newcommand {\Unt} {{\cal U}}
\newcommand {\Leq} {\leq}
\newcommand {\Inf} {\infty}

\newcommand{\hide}[1]{}

\newtheorem{theorem}{Theorem}

\newtheorem{corollary}{Corollary}
\newtheorem{proposition}{Proposition}
\newtheorem{definition}{Definition}
\newtheorem{example}{Example}

 \title[Theory and Practice of Logic Programming]
 { Reasoning about Actions \\
 with Temporal Answer Sets
}

\author[L. Giordano, A. Martelli, D. Theseider Dupr{\'e} ]
{Laura Giordano \\
Dipartimento di Informatica, Universit\`a del Piemonte Orientale, Italy\\
laura@mfn.unipmn.it \\
\and Alberto Martelli\\ Dipartimento di Informatica, Universit\`a di Torino, Italy\\
mrt@di.unito.it\\
\and Daniele Theseider Dupr{\'e} \\
Dipartimento di Informatica, Universit\`a del Piemonte Orientale, Italy\\
dtd@mfn.unipmn.it \\
}

\submitted {15 November 2010}
 \revised {16 April 2011}
 \accepted{31 August 2011}

\jdate{}
\pubyear{}
\pagerange{\pageref{firstpage}--\pageref{lastpage}}
\doi{}

\begin{document}
\bibliographystyle{acmtrans}

\label{firstpage}

\maketitle
\begin{abstract}
In this paper we combine Answer Set Programming (ASP)  with Dynamic Linear
Time Temporal Logic (DLTL) to define a temporal logic programming language for reasoning about
complex actions and infinite computations.
DLTL extends propositional temporal logic of linear time with regular programs of propositional dynamic logic,
which are used for indexing temporal modalities. The action language allows general DLTL formulas to be  included
in domain descriptions to constrain the space of possible extensions.
We introduce a notion of Temporal Answer Set for domain descriptions,
based on the usual notion of Answer Set. Also, we provide a translation of domain descriptions into
standard ASP and we use Bounded Model Checking techniques
for the verification of DLTL constraints.
\end{abstract}
\begin{keywords}
Answer Set Programming, Temporal Logic, Bounded Model Checking.
\end{keywords}

\section{Introduction}
Temporal logic is one of the main tools used in the verification of dynamic systems.
In the last decades, temporal logic has been widely used also in AI in the context of planning, diagnosis, web service verification,
agent interaction and, in general, in most of those areas having to do with some form of reasoning about actions.

The need of temporally extended goals in the context of planning has been first motivated
in \cite{Bacchus98,Kabanza,Giunchiglia&Traverso99}.
In particular,  \cite{Giunchiglia&Traverso99}
developed the idea of planning as model checking in a temporal logic,  where the properties of planning domains
are formalized as temporal formulas in CTL.
In general, temporal formulas can be usefully exploited both in the specification of a domain and in the verification of its properties.
This has been done, for instance, for modeling the interaction of services on the web \cite{Pistore}, as well as for
the specification and verification of agent communication protocols  \cite{GMS-JAL06}.
Recently, Cla{\ss}en and Lakemeyer \cite{Classen:08} have introduced a second order extension
of the temporal logic CTL*, ${\cal ESG} $,
to express and reason about non-terminating Golog programs. The ability to capture infinite computations is important as
agents and robots usually fulfill non-terminating tasks.

In this paper we combine Answer Set Programming (ASP)  \cite{Gelfond}  with Dynamic Linear
Time Temporal Logic (DLTL) \cite{Henriksen99} to define a temporal logic programming language for reasoning about
complex actions and infinite computations.
DLTL extends propositional temporal logic of linear time with regular programs of propositional dynamic logic,
which are used for indexing temporal modalities.
Allowing program expressions within temporal formulas and including arbitrary temporal formulas in domain descriptions
provides a simple way of constraining the (possibly infinite) evolutions of a system,
as in Propositional Dynamic Logic (PDL).
To combine ASP and DLTL, we define a temporal extension of ASP
by allowing temporal modalities to occur within rules
and we introduce a notion of Temporal Answer Set, which captures the temporal dimension of the language as a linear structure
and naturally allows to deal with infinite computations.
A domain description consists of two parts:
a set of temporal rules (action laws, causal laws,  etc.) and a set of constraints (arbitrary DLTL formulas).
The temporal answer sets of the rules in the domain description which also satisfy the constraints
are defined to be the extensions of the domain description.

We provide a translation into standard ASP for
the temporal rules
of the domain description. The temporal answer sets
of an action theory can then be computed as the standard answer sets of the translation.
To compute the extensions of a domain description,
the temporal constraints are evaluated over temporal answer sets
using {\em bounded model checking} techniques \cite{Cimatti03}.  The approach proposed for the verification
of  DLTL formulas extends
the one developed in \cite{Niemela03} for bounded LTL model checking with Stable Models.

The outline of the paper is as follows. In Section 2, we recall the temporal logic DLTL. In Section 3, we introduce our action theory
in temporal ASP.
In Section 4, we define the notions of temporal answer set and extension of a domain description.
Section 5 describes the reasoning tasks, while Sections 6 and 7 describe the model checking problem and provide
a translation of temporal domain descriptions into ASP. Section 8 is devoted to conclusions and related work.

\section{Dynamic Linear Time Temporal Logic }
\label{sec:DLTL}

In this section we briefly define the syntax and semantics of
DLTL as introduced in \cite{Henriksen99}. In such a linear time
temporal logic the next state modality is indexed by actions.
Moreover (and this is the extension to LTL), the until operator $ \Unt^\pi$
is indexed by a program $\pi$ as in PDL.
In addition to the usual  $\Box$ (always) and $\Diamond$ (eventually) temporal modalities
of LTL, new modalities $[\pi]$ and  $\la \pi \ra$ are allowed.
Informally, a formula $[\pi]  \alpha$ is true in a world $w$ of a linear temporal model
(a sequence of propositional interpretations)
if $\alpha$ holds in all the worlds of the model which are reachable from $w$
through any execution of the program $\pi$. A formula $\la \pi \ra  \alpha$ is true in a world $w$ of a linear temporal model
if there exists a world of the model reachable from $w$
through an execution of the program $\pi$, in which $\alpha$ holds.
The program $\pi$ can be any regular expression built from atomic actions using
sequence ($;$), nondeterministic choice ($+$) and finite iteration ($*$).
The usual
modalities $\Box$, $\Diamond $  and $\bigcirc$ (next) of LTL are definable.

Let $\Sigma$ be a finite non-empty alphabet representing actions. Let $\Sigma^*$ and $\Sigma^\omega$ be the
set of finite and infinite words on $\Sigma$,
and let $\Sigma^\Inf$ =$\Sigma^* \cup
\Sigma^\omega$. We denote by $\sigma, \sigma'$ the words over
$\Sigma^\omega$ and by $\tau, \tau'$ the words over $\Sigma^*$.
For $u\in \Sigma^\Inf$,
we denote by {\em prf(u)} the set of finite prefixes of $u$.
Moreover, we denote by $\Leq$ the usual prefix ordering over
$\Sigma^*$
namely, $\tau \leq
\tau'$ iff $\exists \tau''$ such that $\tau\tau'' = \tau'$,
and $\tau < \tau'$ iff $\tau \leq \tau'$ and $\tau \not =
\tau'$.

The set of programs (regular expressions) $Prg(\Sigma)$
generated by $\Sigma$ is:
\begin{center}
$Prg(\Sigma)$ ::= $a$ $\mid$ $\pi_1 + \pi_2$ $\mid$  $\pi_1 ;
\pi_2$ $\mid$ $\pi^*$,
\end{center}
where $a \in \Sigma$ and $\pi_1,\pi_2,\pi$ range over
$Prg(\Sigma)$. A set of finite words is associated with each
program by the mapping $[[]]: Prg(\Sigma) \rightarrow
2^{\Sigma^*}$, which is defined
as follows:
\begin{itemize}
\item
$[[a]]=\{a\}$;
\item $[[\pi_1 + \pi_2]]= [[\pi_1]] \cup [[\pi_2]]$;
\item $[[\pi_1 ; \pi_2]]=\{ \tau_1 \tau_2 \mid   \tau_1 \in [[\pi_1]]
\mbox{ and }  \tau_2 \in [[\pi_2]] \}$;
\item $[[\pi^* ]]= \bigcup [[\pi^i]]$, where
\begin{itemize}
\item $[[\pi^0 ]]= \{ \varepsilon \}$
\item $[[\pi^{i+1} ]]=
    \{ \tau_1 \tau_2 \mid   \tau_1 \in [[\pi]]
    \mbox{ and }  \tau_2 \in [[\pi^i]] \}$,
    for every $i\in \omega$
\end{itemize}
\end{itemize}
where $\varepsilon$ is the empty word (the empty action sequence).

 Let ${\cal P}= \{p_1,p_2,\ldots\}$ be a countable set of
atomic
propositions containing $\top$ and $\bot$ (standing for {\em true} and {\em false}),
and let
DLTL($\Sigma$) ::= $p$ $\mid$ $\neg \alpha$ $\mid$  $\alpha \vee
\beta$ $\mid$ $\alpha \Unt^\pi \beta$,
where $p \in {\cal P} $ and $\alpha, \beta$ range over
DLTL($\Sigma$).

A model of DLTL($\Sigma$) is a pair $M=(\sigma, V)$ where $\sigma
\in \Sigma^\omega$ and $V:\textit{prf}(\sigma) \rightarrow 2^{\cal P}$ is a
valuation function. Given a model $M=(\sigma, V)$, a finite word
$\tau \in \textit{prf}(\sigma)$ and a formula $\alpha$, the satisfiability
of a formula $\alpha$ at $\tau$ in $M$, written $M, \tau \models
\alpha$, is defined as follows:
\begin{itemize}
\item
$M, \tau \models \top$;

\item
$M, \tau \not \models \bot$;

\item
$M, \tau \models p$ iff $p \in V(\tau)$;

\item
$M, \tau \models \neg \alpha$ iff $M, \tau \not \models \alpha$;

\item
$ M, \tau \models \alpha \vee \beta$ iff $M, \tau \models \alpha$
or $M, \tau \models \beta$;

\item
$M, \tau \models  \alpha \Unt^\pi \beta$ iff there exists
$\tau'\in [[\pi]]$ such that $\tau \tau' \in \textit{prf}(\sigma)$ and $M,
\tau \tau' \models \beta$. Moreover, for every $\tau''$ such that
$\varepsilon \Leq \tau'' < \tau'$,  $M, \tau \tau'' \models \alpha$.
\end{itemize}
A formula $\alpha$ is satisfiable iff there is a model $M=(\sigma,
V)$ and a finite word $\tau \in \textit{prf}(\sigma)$ such that $M, \tau
\models  \alpha$.
The formula $\alpha \Unt^\pi \beta$  is true at $\tau$ if
``$\alpha$ until $\beta$''  is true on a finite stretch of
behavior which is in the linear time behavior of the program
$\pi$.

The classical connectives $\supset$ and $\wedge$ are defined as usual.
The derived modalities $\la \pi \ra$ and $[\pi]$ can be defined as
follows: $\la \pi \ra \alpha \equiv \top \Unt^\pi \alpha$ and
$[\pi] \alpha \equiv \neg \la \pi \ra \neg \alpha$.
Furthermore, if we let $\Sigma= \{a_1,\ldots,a_n\}$, the $\Unt$ (until),
$\bigcirc$ (next), $\Diamond$ and $\Box$ operators of LTL can be defined
as follows: $\bigcirc \alpha \equiv \bigvee_{a \in \Sigma} \la a
\ra \alpha$,\ \ $\alpha \Unt \beta \equiv \alpha \Unt^{\Sigma^*}
\beta$, \ \ $\Diamond \alpha \equiv \top \Unt \alpha$, \ \ $\Box
\alpha \equiv \neg \Diamond \neg \alpha$,\ \ where, in
$\Unt^{\Sigma^*}$, $\Sigma$ is taken to be a shorthand for the
program $a_1 + \ldots +a_n$. Hence, LTL($\Sigma$)
is a fragment of DLTL($\Sigma$). As shown in \cite{Henriksen99},
DLTL($\Sigma$) is strictly more expressive than LTL($\Sigma$). In
fact, DLTL has the full expressive power of the monadic second
order theory of $\omega$-sequences.

\section{Action theories in Temporal ASP} \label{sec:actiontheories}
\label{sec:ActionTheory}

Let ${\cal P}$ be a set of atomic propositions, the {\em fluent
names}.
A {\em simple fluent literal} $l$ is a fluent name  $f$ or its
negation $\neg f$. Given a fluent literal $l$, such that $l=f$ or
$l= \neg f$, we define $|l|=f$. We denote by ${Lit_S}$ the set
of all simple fluent literals and, for each $l \in Lit_S$, we denote by $\overline{l}$
the complementary literal  (namely,  $\overline{p}= \neg p$ and  $\overline{\neg p}= p$).
$Lit_T$ is the set of {\em temporal fluent literals}:
if $l \in Lit_S$, then $[a]l, \bigcirc l \in Lit_T$ (for $a\in \Sigma$).
Let $Lit = Lit_S \cup Lit_T \cup \{\bot\}$.
Given a (temporal) fluent literal $l$, $not\;l$ represents the default negation of $l$.
A (temporal) fluent literal, possibly preceded by a default negation, will be called an
 {\em extended fluent literal}.

A {\em state} is a set of fluent literals in ${Lit_S}$.
A state is {\em consistent} if it is not the case that both $f$ and $\neg f$ belong to the state, or that $\bot$ belongs to the state.
A state is {\em complete} if, for each fluent name $p \in \cal{P}$, either $p$ or $\neg p$ belongs to it.
The execution of an action in a state may possibly change the values of fluents in the state through its direct and indirect effects,
thus giving rise to a new state.

Given a set of actions $\Sigma$, a {\em domain description $D$ over $\Sigma$} is defined as a tuple
$(\Pi, {\cal C})$, where $\Pi$  is a set of laws
({\em action laws},  {\em causal laws}, {\em precondition laws}, etc.)
describing the preconditions and effects of actions,
and  ${\cal C}$ is a set of {\em
DLTL constraints}.
While $\Pi$ contains the laws that are usually included in a domain description,
which define the executability conditions for actions, their direct and indirect effects as well as
conditions on the initial state,
${\cal C}$ contains general DLTL constraints which must be satisfied by the
intended interpretations of the domain description.
While the laws in $\Pi$ define conditions
on single states or on pairs of consecutive states,
DLTL constraints define more general conditions on possible sequences of states and actions.
Let us first describe the laws occurring in $\Pi$.

The {\bf action laws} describe the immediate effects of actions. They have the form:
\begin{equation} \label{a_law}
\Box ( [a] l_0  \leftarrow t_{1}, \ldots ,t_m , not\; t_{m+1} , \ldots ,  not\; t_n)
 \end{equation}
where $l_0$ is a simple fluent literal and the $t_i$'s are either simple fluent literals or temporal fluent literals of the form $[a]l$.
 Its meaning is that executing action
$a$ in a state in which the conditions $t_{1}, \ldots ,t_m$ hold and conditions $ t_{m+1}, \ldots , t_n$ do not hold
causes the effect $l_0$ to hold.
Observe that a temporal literal $[a]l$ is true in a state when the execution of action $a$ in that state
causes $l$ to become true in the next state.
For instance,
the following action laws describe the deterministic effect of the actions {\em shoot} and {\em load} for the Russian Turkey problem:
\begin{center}
$\Box( [shoot] \neg alive \leftarrow loaded)$\\
$\Box [load] loaded$
\end{center}

Non deterministic actions can be defined using default negation in the body of action laws.
In the example,
after spinning the gun, it may be loaded or not:
\begin{center}
$\Box( [spin] loaded \leftarrow \; not \; [spin] \neg loaded) $\\
$\Box( [spin] \neg loaded  \leftarrow \; not \; [spin] loaded) $
\end{center}
Observe that, in this case, temporal fluent literals occur in the body of action laws.

{\bf Causal laws} are intended to express ``causal" dependencies among fluents.
In $\Pi$ we allow two kinds of causal laws.
{\bf  Static causal laws} have the form:
\begin{equation} \label{sc_law}
\Box ( l_0 \leftarrow l_{1}, \ldots ,l_m , not\; l_{m+1} , \ldots ,  not\; l_n)
\end{equation}
where the $l_i$'s  are simple fluent literals.
Their meaning is: if $ l_{1}, \ldots , l_m$ hold in a state
and $ l_{m+1}, \ldots , l_n$ do not hold in that state,
than  $l_0$ is caused to hold in that state.

{\bf Dynamic causal laws} have the form:
\begin{equation} \label{dc_law}
\Box ( \bigcirc l_0  \leftarrow t_{1}, \ldots ,t_m , not\; t_{m
+1} , \ldots , not\; t_n)
\end{equation}
where $l_0$ is a simple fluent literal and the $t_i$'s are either simple fluent literals or temporal fluent literals
of the form $\bigcirc l_i$.
Their meaning is: if $ t_{1}, \ldots , t_m$ hold
and $ t_{m+1}, \ldots , t_n$ do not hold in a state,
then $l_0$ is caused to hold in the next state.
Observe that $t_i= \bigcirc l_i$ holds in a state when $l_i$ holds in the next state.

For instance, the static causal law $\Box(  frightened \leftarrow in\_sight , alive)$
states that the turkey being in sight of the hunter
causes it to be frightened, if it is alive; alternatively, the dynamic causal law
$\Box(  \bigcirc frightened \leftarrow \bigcirc in\_sight , \neg in\_sight , alive)$
states that if the turkey is alive, it {\em becomes} frightened (if it is not already) when it {\em starts} seeing the hunter;
but it can possibly become non-frightened later, due to other events,
while still being in sight of the hunter\footnote{Shorthands like those in \cite{Denecker98} could be used, even though we do not introduce them in this paper, to express that a fluent or a complex formula is initiated (i.e., it
is false in the current state and caused true in the next one).}.

Besides action laws and causal laws, that apply to all states,
we also allow for laws in $\Pi$ that only apply to the initial state.
They are called  {\bf initial state laws} and have the form:
\begin{equation} \label{is_law}
l_0  \leftarrow l_{1}, \ldots , l_m , not\; l_{m+1} , \ldots ,  not\; l_n
\end{equation}
where the $l_i$'s are simple fluent literals.
Observe that initial state laws, unlike static causal laws, only apply to the initial state
as they are not prefixed by the $\Box$ modality.
As a special case, the initial state can be defined as a set of simple fluent literals. For instance,
the initial state
$\{ alive, \neg  in\_sight, \neg frightened \}$ is defined by
the initial state laws:
$$alive \; \; \; \; \; \;  \neg  in\_sight  \; \; \; \; \; \; \neg frightened $$

Given the laws introduced above,
all the usual ingredients of action theories
can be introduced in $\Pi$.
In particular, let us consider the case when $\bot$ can occur as a literal in the head of those laws.

{\bf Precondition laws} are special kinds of action laws (\ref{a_law}) with $\bot$ as effect. They have the form:
$$\Box ( [a]\bottom \leftarrow l_1, \ldots ,l_m , not\; l_{m+1} , \ldots , not\; l_n) $$
where $a\in \Sigma$ and  the $l_i$'s are simple fluent literals.
The meaning is that the execution of an action $a$ is not
possible in a state in which  $l_1,\ldots,l_{m}$ hold  and $ l_{m+1}, \ldots , l_n$ do not hold (that is,  no state may
result from the execution of $a$ in a state in which $l_1,\ldots,l_m$ hold and $ l_{m+1}, \ldots , l_n$ do not hold).

{\bf State constraints} that apply to the initial state or to all states can be obtained
when  $\bot$ occurs
in the head of initial state laws (\ref{is_law})
or static causal laws (\ref{sc_law}):
\begin{center}
$ \bot  \leftarrow l_{1}, \ldots ,l_m , not\; l_{m+1} , \ldots ,  not\; l_n $ \\
$\Box (\bot \leftarrow l_{1}, \ldots ,l_m , not\; l_{m+1} , \ldots ,  not\; l_n) $
\end{center}
The first one means that it is not the case that, in the initial state,  $l_1,\ldots,l_{m}$ hold  and $ l_{m+1}, \ldots , l_n$ do not hold.
The second one means that there is no state in which  $l_1,\ldots,l_{m}$ hold  and $ l_{m+1}, \ldots , l_n$ do not hold.

As in \cite{Lifschitz:90} we call {\em frame} fluents
those fluents to which the law of inertia applies.
The persistency of frame fluents from a state to the next one
can be enforced by introducing in $\Pi$ a set of laws, called {\bf persistency laws},
\begin{center}
$\Box (\bigcirc f \leftarrow  f , \; not \bigcirc\neg f)$\\
$\Box (\bigcirc \neg f \leftarrow  \neg f , \; not \bigcirc f)$
\end{center}
for each simple fluent $f$ to which inertia applies.
Their meaning is that, if $f$ holds in a state,
then $f$ will hold in the next state, unless its complement  $\neg f$ is caused to hold. And similarly for $\neg f$.
Note that persistency laws are instances of dynamic causal laws (\ref{dc_law}).
In the following, we use {\bf inertial f} as a shorthand for the persistency laws for $f$.

The persistency of a fluent from a state to the next one is blocked by the execution of an action which causes the value of the fluent to change, or by a nondeterministic action which may cause it to change.
For instance, the persistency of $\neg loaded$ is blocked by $load$ and by $spin$.

Examples of non-inertial fluents, for which persistency laws are not included,
are those taking a default truth value, as for
a spring door which is normally closed:
$$\Box (closed \leftarrow  not \neg closed)$$
or those which always change, at least by default, e.g., in case of a
pendulum (see \cite{Giunchiglia&al:2004}) always switching between left and right position:
\begin{center}
$\Box (\bigcirc right \leftarrow  \neg right , \; not \bigcirc \neg right)$\\
$\Box (\bigcirc  \neg right \leftarrow  right , \; not \bigcirc right)$
\end{center}
Such default action laws play a role similar to that of {\em inertia rules} in ${\cal C}$ \cite{Giunchiglia&al:98},
${\cal C}^+$ \cite{Giunchiglia&al:2004}
and  ${\cal K}$ \cite{Leone04}.

Initial state laws may incompletely specify the initial state.
In this paper we want to reason about complete states so that the execution of an infinite sequence of actions
gives rise to a linear model as defined in section 2. For this reason,
we assume that,  for each fluent $f$, $\Pi$ contains the laws:
\begin{center}\label{complete_state_law}
$f\  \leftarrow \; not \; \neg f$ \\
$\neg f \leftarrow   \; not \;  f$
\end{center}
As we will see later, this assumption in general is not sufficient to guarantee that all the states are complete.

{\em Test actions}, useful for checking the value of a fluent in a state in the definition of complex actions, can be defined through suitable laws as follows.
Given a simple fluent literal $l \in Lit_S$, the test action $l?$
is executable only if $l$ holds, and it has no effect on any fluent $f$:
\begin{center}
$\Box( [l?] \bot \leftarrow not \; l)$ \\
$\Box( [l?] f \leftarrow  f)$ \\
$\Box( [l?] \neg f \leftarrow  \neg f)$
\end{center}

The second component of a domain description
is the set ${\cal C}$ of {\bf DLTL constraints},
which allow very general temporal conditions to be imposed on the executions of the domain description
(we will call them extensions).
Their effect is that of restricting the space of the possible executions.
For instance,
the constraint:
$$\neg loaded \; \Unt \; in\_sight$$
states that
the gun is not loaded until the turkey is in sight.
Its addition filters out all the executions
in which the gun is loaded before the turkey is in sight.

A temporal constraint can also require a complex behavior to be performed.
The program
 \begin{equation}\label{program}
 (\neg in\_sight?; wait)^* ;in\_sight?; load;shoot
 \end{equation}
describes the behavior of the hunter who waits for a turkey until
it appears and, when it is in sight, loads the gun and shoots.
Actions  $in\_sight?$ and $\neg in\_sight?$ are test actions, as introduced before.
If the constraint
 $$\la (\neg in\_sight?; wait)^* ;in\_sight?; load;shoot \ra \top$$
 is included in ${\cal C}$
then all the runs of the domain description which do not start with an execution of the given program will be
filtered out.  For instance, an extension in which in the initial state the turkey is not in sight  and
the hunter loads the gun and shoots is not allowed.
In general, the inclusion of a constraint  $\la \pi \ra \top$ in ${\cal C}$ requires that there is an execution
of the program $\pi$ starting from the initial state.

\begin{example} \label{es_shooting}
Let us consider a variant of the Yale shooting problem
including some of the laws above, and
some more stating that: if the hunter is in sight and the turkey is alive, the turkey is frightened;
the turkey may come in sight or out of sight (nondeterministically) during waiting.

Let $\Sigma =\{ load, shoot, spin, wait \}$ and ${\cal P}= \{alive, loaded, in\_sight, frightened\} $.
We define a domain description ($\Pi$,${\cal C}$), where
$\Pi$ contains the following laws:

Immediate effects:

\begin{tabbing}

$\Box( [wait] in\_sight \leftarrow not [wait] \neg in\_sight ) $ \=  \kill

$\Box( [shoot] \neg alive \leftarrow loaded)$ \>

$\Box [load] loaded$ \\

$\Box( [spin] loaded   \leftarrow not [spin] \neg loaded) $ \>

$\Box( [spin] \neg  loaded   \leftarrow not [spin] loaded) $ \\

$\Box( [wait] in\_sight \leftarrow  not [wait] \neg in\_sight ) $ \>

$\Box( [wait] \neg  in\_sight \leftarrow not [wait] in\_sight ) $

\end{tabbing}

Causal laws: \ \ \ \ \ \ \ \ \ \ \
$\Box(  frightened \leftarrow in\_sight , alive)$

Initial state laws: \ \ \ \ \
$alive$ \ \ \ \ \ \ \
$\neg  in\_sight$  \ \ \ \ \ \ \
$\neg frightened $

Precondition laws: \ \ \
 $\Box ( [load]\bottom \leftarrow loaded) $

\noindent
All fluents in ${\cal P}$ are inertial: {\bf inertial} alive,  {\bf inertial} loaded,  {\bf inertial} in\_sight,  {\bf inertial} frightened;
and
${\cal C}=\{ \neg loaded \; \Unt \; in\_sight\}$.

 Given this domain description we may want to ask if it is possible for the hunter to execute
 a behavior described by program $\pi$ in (\ref{program})
 so that the turkey is not alive after that execution.
The intended answer to the query $\la \pi \ra \neg alive$ would be yes, since there is a possible scenario in which this can happen.
\end{example}
\begin{example} \label{es_mailbox}
In order to see that the action theory in this paper is well suited to deal with infinite executions,
consider a mail delivery agent, which repeatedly checks if there is mail in the mailboxes of $a$ and $b$
and then it delivers the mail to $a$ or to $b$, if there is any; otherwise, it waits.
Then, the agent starts again the cycle.
The actions in $\Sigma$ are:
$begin$, $sense\_mail(a)$ (the agent verifies if there is mail in the mailbox of $a$), $sense\_mail(b)$,
$deliver(a)$ (the agent delivers the mail to $a$), $deliver(b)$, $wait$ (the agent waits).
The fluent names are $mail(a)$ (there is mail in the mailbox of $a$)  and $mail(b)$.
The domain description contains the following laws for $a$:

Immediate effects:
\begin{quote}
$\Box [deliver(a)] \neg mail(a)$

$\Box( [sense\_mail(a)] mail(a)   \leftarrow \; not \; [sense\_mail(a)] \neg mail(a))$
\end{quote}

Precondition laws:
\begin{quote}
 $\Box ( [deliver(a)]\bottom \leftarrow \neg mail(a)) $

 $\Box ( [wait]\bottom \leftarrow mail(a)) $
\end{quote}
Their meaning is (in the order) that: after delivering mail to $a$, there is no mail for $a$ anymore;  the action $sense\_mail(a)$ of
verifying if there is mail for $a$,
may (non-monotonically) cause $mail(a)$ to become true;
in case there is no mail for $a$, $deliver(a)$ is not executable;
in case there is mail for $a$, wait is not executable.
The same laws are also introduced for the actions involving $b$.

\noindent
All fluents in ${\cal P}$ are inertial:
 {\bf inertial} mail(a),  {\bf inertial} mail(b).
Observe that, the persistency laws for inertial fluents interact with the immediate effect laws above.
The execution of $sense\_mail(a)$
in a state in which there is no mail for $a$ ($\neg mail(a)$), may either lead to a state in which $mail(a)$ holds (by the second action law) or to a state in which  $\neg mail(a)$ holds (by the persistency of $\neg mail(a)$).

${\cal C}$ contains the following constraints:
\begin{quote}
$\la begin \ra \top$

$\Box [begin] \la sense(a); sense(b); (deliver(a)+ deliver(b) + wait ); begin \ra \top $

\end{quote}
The first one means that the action $begin$ must be executed in the initial state.
The second one means that, after any execution of action $begin$,
the agent must execute $sense(a)$ and $sense(b)$ in the order, then either deliver the mail to $a$ or to $b$ or wait and,
then, execute action $begin$ again, to start a new cycle.

We may want to check that if there is mail for $b$, the agent will eventually deliver it to $b$.
This property, which can be formalized by the formula
$\Box (mail(b) \supset \Diamond \neg mail(b))$,
does not hold as there is a possible scenario in which there is mail for $b$,
but the mail is repeatedly delivered to $a$ and never to $b$.
The mail delivery agent we have described is not fair.
\end{example}

\begin{example}
\label{es_sistema}
As an example of modeling a controlled system and its possible faults,
we describe an adaptation of the qualitative causal model
of the ``common rail'' diesel injection system from \cite{Panati01} where:
\begin{itemize}
   \item
   Pressurized fuel is stored in a container, the {\em rail}, in order to be injected at high pressure into the cylinders.
   We ignore in the model the output flow through the injectors.
   Fuel from the tank is input to the rail through a {\em pump}.
   \item
   A regulating system, including, in the physical system, a {\em pressure sensor}, a {\em pressure regulator} and an {\em Electronic Control Unit}, controls pressure in the rail; in particular, the pressure regulator, commanded by the ECU based on the measured pressure, outputs fuel back to the tank.
   \item
   The control system repeatedly executes the \textit{sense\_p} (sense pressure) action while the physical system evolves through internal events.
\end{itemize}

\noindent
Examples of formulas from the model are contained in $\Pi$:
\begin{quote}
$\Box( [pump\_weak\_fault]  f\_in\_low)$
\end{quote}
shows the effect of the fault event $pump\_weak\_fault$.
Flows influence the derivative of the pressure in the rail, and the pressure derivative influences pressure via the
event $p\_change$:
\begin{small}
\begin{tabbing}
$\Box( p\_incr   \leftarrow  f\_out\_very\_low , f\_in\_low ) $ \=  \kill
$\Box( p\_decr   \leftarrow f\_out\_ok , f\_in\_low ) $ \>
$\Box( p\_incr   \leftarrow  f\_out\_very\_low , f\_in\_low ) $\\
$\Box( p\_steady   \leftarrow f\_out\_low , f\_in\_low ) $ \>
$\Box( [p\_change] p\_low \leftarrow p\_ok , p\_decr)$\\
$\Box( [p\_change] p\_ok \leftarrow p\_low , p\_incr)$\>
$\Box( [p\_change] \bot \leftarrow p\_steady ) $\\
$\Box( [p\_change] \bot \leftarrow p\_decr , p\_low ) $\>
$\Box( [p\_change] \bot \leftarrow p\_incr , p\_ok ) $
\end{tabbing}
\end{small}
\noindent
The model of the pressure regulating subsystem includes:
\begin{small}
\begin{tabbing}
$\Box( [sense\_p] p\_obs\_ok \leftarrow p\_ok)$ \= $BLA BLA$ \= \kill
$\Box( [sense\_p] p\_obs\_ok \leftarrow p\_ok)$ \> \>
$\Box( f\_out\_ok \leftarrow normal\_mode , p\_obs\_ok ) $\\
$\Box( [sense\_p] p\_obs\_low \leftarrow p\_low)$ \> \>
$\Box( f\_out\_low \leftarrow comp\_mode , p\_obs\_ok ) $ \\
$\Box( [switch\_mode]  comp\_mode)$\> \>
$\Box( f\_out\_very\_low \leftarrow comp\_mode , p\_obs\_low ) $
\end{tabbing}
\end{small}
\noindent
with the obvious mutual exclusion constraints among fluents.
Initially, everything is normal and pressure is steady:
$ p\_ok$,
$ p\_steady$,
$ f\_in\_ok $,
$ f\_out\_ok$,
$normal\_mode$.

All fluents are inertial. We have the following
temporal constraints in ${\cal C}$:
\begin{quote}
\noindent
$\Box( (p\_ok \wedge p\_decr) \orr (p\_low \wedge p\_incr) \supset  \la p\_change \ra \top)$\\
$\Box( normal\_mode \wedge p\_obs\_low \supset  \la switch\_mode \ra  \top) $\\
$ [sense\_p] \la(\Sigma - \{ sense\_p\})^* \ra \la sense\_p\ra \top $\\
$\Box[pump\_weak\_fault]  \neg \Diamond \la pump\_weak\_fault \ra \top $
\end{quote}
The first one models conditions which imply a pressure change.
The second one models the fact that a mode switch occurs when the system is operating in normal mode and
the measured pressure is low.
The third one models the fact that the control system repeatedly executes $sense\_p$, but other actions may occur in between.
The fourth one imposes that at most one fault may occur in a run.

Given this specification we can, for instance, check that if pressure is low in one state, it will be normal in the third next one,
namely, that the temporal formula $\Box (p\_low \supset \bigcirc\bigcirc\bigcirc p\_ok)$ is satisfied in all the possible scenarios
admitted by the domain description.
That is, the system tolerates a weak fault of the pump --- the only fault included in this model.
In general, we could, e.g., be interested in proving properties that hold if at most one fault occurs, or at most one fault
in a set of ``weak'' faults occurs.
\end{example}

As we have seen from the examples,
our formalism allows naturally to deal with infinite executions of actions.
Such infinite executions define the models over which temporal formulas can be evaluated.
In order to deal with cases (e.g., in planning) where we want to reason on finite action sequences,
it is easy to see that any finite action sequence can be represented as an infinite one,
adding to the domain description an action
\emph{dummy}, and the constraints
$\Diamond \la dummy \ra \top $ and
$\Box[dummy] \la dummy \ra \top $
stating that action \emph{dummy} is eventually executed and, from that point on, only the action \emph{dummy} is executed.
In the following, we will restrict our attention to infinite executions,  assuming that the dummy
action is introduced when needed.

	\section{Temporal answer sets and extensions for domain descriptions}
\label{sec:extensions}

Given a domain description $D= (\Pi,{\cal C})$,
the laws in $\Pi$
are rules of a general logic program extended
with a restricted use of temporal modalities.
In order to define the extensions of a domain description,
we introduce a notion of {\em temporal answer set}, extending the usual notion
of {\em answer set} \cite{Gelfond}.
The extensions of a domain description will then be defined as the temporal answer sets of
$\Pi$ satisfying the integrity constraints ${\cal C}$.

In the following, for conciseness, we call ``simple (temporal) literals'' the ``simple (temporal) fluent literals".
We call {\em rules}
the laws in $\Pi$, having one of the two forms:
\begin{equation}  \label{rule1}
 l_0 \leftarrow l_1, \ldots , l_m , not\; l_{m+1} , \ldots , not\; l_n
\end{equation}
where the $l'_i$'s are simple literals, and
\begin{equation} \label{rule2}
\Box (  t_0 \leftarrow t_1, \ldots , t_m , not\; t_{m+1} , \ldots , not\; t_n )
\end{equation}
where the $t_i$'s are simple or temporal literals, the first one capturing initial state laws,
the second one all the other laws.
To define the notion of extension, we also need to introduce rules of the form:
$[a_1;\ldots;a_h] ( t_0 \leftarrow t_1, \ldots , t_m, not\; t_{m+1} , \ldots , not\; t_n) $,
where the $t_i$'s are simple or temporal literals, which will be used to define the reduct of a program.
The modality $[a_1;\ldots;a_h]$ means that the rule applies in the state obtained after the execution of actions $a_1,\ldots, a_h$.
Conveniently, also the notion of temporal literal used so far needs to be extended to include literals of the form $[a_1;\ldots;a_h] l$,
meaning that
$l$ holds after
the action sequence $a_1,\ldots, a_h$.

As we have seen, temporal models of DLTL are linear models, consisting of an action sequence $\sigma$ and a valuation
function $V$, associating  a propositional evaluation with each state in the sequence (denoted by a prefix of $\sigma$).
We extend  the notion of answer set to capture this  linear structure of temporal models,
by defining a partial temporal interpretation  as a pair $(\sigma, S)$, where $\sigma \in \Sigma ^\omega$ and $S$ is
a set of literals of the form  $[a_1;\ldots;a_k]l$,
where $a_1\ldots  a_k$ is a prefix of $\sigma$.

\begin{definition}
Let $\sigma \in \Sigma ^\omega$. A {\em partial temporal interpretation over $\sigma$} is a pair $(\sigma,S)$
where $S$ is a set of temporal literals of the
form $ [a_1;\ldots;a_k]l$,
such that $a_1\ldots a_k$ is a prefix of $\sigma$, and it is not the case that both $ [a_1;\ldots;a_k]l$ and $ [a_1;\ldots;a_k]\neg l$
belong to $S$ or that $ [a_1;\ldots;a_k]\bot$ belongs to $S$ (namely, $S$ is a  {\em consistent}  set of temporal literals).
\end{definition}
A temporal interpretation $(\sigma,S)$ is said to be {\em total} if either $ [a_1;\ldots;a_k]p\in S$ or $ [a_1;\ldots;a_k]\neg p \in S$,
for each $a_1\ldots a_k$ prefix of $\sigma$ and for each fluent name $p$.

Observe that a partial interpretation $(\sigma,S)$ provides, for each prefix $a_1\ldots a_k$, a partial evaluation of fluents in the
state corresponding to that prefix.
The (partial) state $w^{(\sigma,S)}_{a_1\ldots a_k}$ obtained by the execution of the actions $a_1\ldots a_k$ in the sequence
can be defined as follows:
\vspace{-0.2cm}
$$w^{(\sigma,S)}_{a_1\ldots a_k} = \{ l :  [a_1;\ldots;a_k]l \in S\}$$
Given a partial temporal interpretation $(\sigma,S)$ and a prefix $a_1\ldots a_k$ of $\sigma$,
we define the {\em satisfiability of a simple, temporal and extended literal $t$
in $(\sigma,S)$ at $a_1\ldots a_k$}
(written $(\sigma,S), a_1\ldots a_k \models t$) as follows:

$(\sigma,S), a_1\ldots a_k \models \top$

$(\sigma,S), a_1\ldots a_k \not \models \bot$

$(\sigma,S), a_1\ldots a_k \models l$ \  \emph{iff} \  $ [a_1;\ldots;a_k]l \in S$, for a simple literal $l $

$(\sigma,S), a_1\ldots a_k \models [a] l$  \  \emph{iff}\   $ [a_1;\ldots;a_k;a]l \in S$ or $a_1\ldots a_k, a$ is not a prefix of $\sigma$

$(\sigma,S), a_1\ldots a_k \models \bigcirc l$  \  \emph{iff}\  $ [a_1;\ldots;a_k;b]l \in S$, where $a_1\ldots a_k b$ is a prefix of
$\sigma$

$(\sigma,S), a_1\ldots a_k \models not\; l$ \  \emph{iff} \  $(\sigma,S), a_1\ldots a_k \not \models l$

\noindent
The satisfiability of rule bodies
in a partial interpretation is defined as usual:

$(\sigma,S), a_1\ldots a_k \models t_1, \ldots, t_n$  \ \emph{iff} \  $(\sigma,S), a_1\ldots a_k \models t_i$ for $i = 1, \ldots, n$.

\noindent
A rule $H \leftarrow Body$ is satisfied in a partial temporal  interpretation $(\sigma,S)$ if,  $(\sigma,S), \varepsilon \models Body$
implies $(\sigma,S), \varepsilon \models H$, where $\varepsilon$ is the empty action sequence.

A rule $\Box (H \leftarrow Body)$ is satisfied in a partial temporal  interpretation $(\sigma,S)$ if,
for all action sequences $a_1\ldots a_k$ (including the empty one),   $(\sigma,S), a_1\ldots a_k \models Body$
implies $(\sigma,S), a_1\ldots a_k \models H$.

A rule $[a_1;\ldots;a_h] ( H \leftarrow Body ) $
is satisfied in a partial temporal  interpretation $(\sigma,S)$ if
$(\sigma,S), a_1\ldots a_h \models Body$
implies $(\sigma,S), a_1\ldots a_h \models H$.

We are now ready to define the notion of answer set for a set $P$ of rules
that does not contain default negation.
Let $P$ be a set of rules over an action alphabet $\Sigma$, not containing default negation, and
let $\sigma \in \Sigma^\omega$.

\begin{definition}
A partial temporal interpretation $(\sigma,S)$ is a {\em temporal answer set of $P$} if
$S$ is minimal (in the sense of set inclusion) among the $S'$ such that $(\sigma,S')$ is a partial interpretation
satisfying the rules in $P$.

\end{definition}

In order to define answer sets of a program $P$ possibly containing negation,
given a partial temporal interpretation $(\sigma, S)$ over
$\sigma \in \Sigma^\omega$, we define the {\em reduct, $P^{(\sigma, S)}$, of  $P$ relative to $(\sigma, S)$} extending
Gelfond and Lifschitz'
transform \cite{Gelfond&Lifschitz:98}
to compute a different reduct of $P$ for each
prefix $a_1,\ldots,a_h$ of $\sigma$.

\begin{definition}
The {\em reduct, $P^{(\sigma, S)}_{a_1,\ldots,a_h}$, of  $P$ relative to $(\sigma, S)$ and to the prefix $a_1,\ldots,a_h$ of $\sigma$ }
is the set of all the rules
$$[a_1;\ldots;a_h] ( H \leftarrow t_1, \ldots , t_m ) $$
such that
$\Box (  H \leftarrow t_1, \ldots , t_m , not\; t_{m+1} , \ldots , not\; t_n )$
is in $P$ and  $(\sigma, S), a_1,\ldots,a_h \not \models  t_i$, for all $i=m+1,\ldots,n$.
The {\em reduct $P^{(\sigma, S)}$ of  $P$ relative to $(\sigma, S)$} is the union of all reducts  $P^{(\sigma, S)}_{a_1,\ldots,a_h}$
 for all prefixes $a_1,\ldots,a_h$ of $\sigma$.

\end{definition}

In essence, given
$(\sigma, S)$,
a different reduct is defined for each finite prefix of $\sigma$,
i.e., for each possible state corresponding to a prefix of $\sigma$.

\begin{definition}
A partial temporal interpretation $(\sigma, S)$ is a {\em temporal answer set of $P$} if
$(\sigma, S)$ is a temporal answer set of the reduct $P^{(\sigma, S)}$.
\end{definition}

The definition above is a natural generalization of the usual notion of answer set to programs with temporal rules.
Observe that $\sigma$ has infinitely many prefixes, so that the reduct $P^{(\sigma, S)}$ is infinite as well as its answer sets.
This is in accordance with the fact that temporal models are infinite.

In the following, we will devote our attention to those domain descriptions $D=(\Pi,\cal C)$
such that $\Pi$ has total temporal answer sets. We will call such domain descriptions {\em well-defined domain descriptions}.
As we will see below, total temporal answer sets can indeed be regarded
as temporal models (according to the definition of model in Section 2).
Although it is not possible to define general syntactic conditions which guarantee that the temporal answer sets of $\Pi$ are total,
this can be done in some specific case.
It is possible to prove the following:

\begin{proposition}\label{total answer sets}
Let $D=(\Pi,\cal C)$  be a domain description  over $\Sigma$, such that all fluents are inertial. Let $\sigma\in \Sigma^\omega$.
Any answer set of $\Pi$ over $\sigma$ is a total answer set over $\sigma$.
\end{proposition}
This result is not surprising, since, as we have assumed in the previous section, the laws for completing the initial state are implicitly added to $\Pi$,
so that the initial state is complete. Moreover,
it can be shown that
(under the conditions, stated in Proposition \ref{total answer sets}, that all fluents are inertial)
the execution of an action in a complete state produces
(nondeterministically,  due to the presence of nondeterministic actions) a new complete state,
which can be only determined by the action laws, causal laws and persistency laws executed in that state.

In the following, we define the notion of {\em extension}
of a well-defined domain description $D=(\Pi,\cal C)$ over $\Sigma$ in two steps:
first, we find the temporal answer sets of $\Pi$;
second, we filter out all the temporal answer sets which do not satisfy the temporal constraints in ${\cal C}$.
For the second step,  we need to define when a temporal formula $\alpha$ is satisfied in a  total temporal interpretation $(\sigma, S)$.
Observe that a total answer set $(\sigma, S)$ can be easily seen as a temporal model, as defined
in Section \ref{sec:DLTL}. Given a total answer set $(\sigma, S)$ we define the corresponding temporal model  as
$M_S=(\sigma,V_S)$, where $p\in V_S(a_1,\ldots,a_h)$ if and only if $[a_1;\ldots;a_h] p \in S$, for all atomic propositions $p$.
We say that a total answer set $S$ over $\sigma$ satisfies  a  DLTL formula $\alpha$  if  $M_S, \varepsilon \models \alpha$.

\begin{definition}
An {\em extension of a well-defined domain domain description $D=(\Pi,\cal C)$ over $\Sigma$} is a (total) answer set $(\sigma, S)$  of $\Pi$  which satisfies
the constraints in ${\cal C}$.
\end{definition}

Notice that, in general, a domain description may have more than one extension
even for the same action sequence $\sigma$:
the different extensions of $D$ with the same $\sigma$
account for the different
possible initial states (when the initial state is incompletely
specified) as well as for the different possible effects of
nondeterministic actions.

\begin{example} \label{es_extension}
Assume the {\em dummy} action is added to the Russian Turkey domain in Section \ref{sec:ActionTheory}.
Given the infinite sequence
$\sigma_1= \neg  in\_sight?;$ $wait;$ $  in\_sight?;$ $ load; shoot; $ $dummy; \ldots$,
the domain description has  (among the others) an extension $(\sigma_1,S_1)$ over $\sigma_1$
containing  the following temporal literals (for the sake of brevity, we write $[a_1; \ldots; a_n] (l_1 \wedge \ldots \wedge l_k)$
 to mean that $[a_1; \ldots; a_n] l_i$ holds in  $S_1$ for all $i$'s):\\
 $[\varepsilon] (alive \wedge \neg in\_sight \wedge  \neg frightened \wedge \neg loaded)$,\\
 $[\neg  in\_sight?] (alive \wedge \neg in\_sight \wedge \neg frightened \wedge \neg loaded)$, \\\
 $[\neg  in\_sight?;$ $ wait] (alive \wedge in\_sight \wedge frightened \wedge \neg loaded)$, \\\
 $[\neg  in\_sight?; wait; in\_sight? ] (alive \wedge in\_sight \wedge frightened \wedge \neg loaded)$, \\
 $[\neg  in\_sight?; wait; in\_sight?; load ] (alive \wedge in\_sight \wedge frightened \wedge loaded)$, \\
 $[\neg  in\_sight?; wait; in\_sight?; load; shoot  ] (\neg alive \wedge  in\_sight \wedge frightened \wedge loaded)$, \\
 $[\neg  in\_sight?; wait; in\_sight?; load; shoot; dummy] (\neg alive \wedge in\_sight \wedge frightened \wedge loaded)$ \\
and so on.
This extension satisfies the constraints in the domain description and
corresponds to a linear temporal model $M_{S_1}=(\sigma_1,V_S)$.
\end{example}

To conclude this section we would like to point out that,
given a  domain description $D=(\Pi,\cal C)$ over $\Sigma$ such that $\Pi$ only admits total answer sets,
a {\em transition system} $(W, I, T)$ can be associated with $\Pi$, as follows:
\begin{itemize}
\item[-]
$W$ is the set of all the possible consistent and complete states of the domain description;

\item[-]
$I$ is the set of all the states in $W$ satisfying the initial state laws in $\Pi$;

\item[-]
$T\subseteq W\times \Sigma \times W $ is the set of all triples $(w,a,w')$ such that:
 $w,w' \in W$, $a \in \Sigma$ and
for some total answer set $(\sigma,S)$ of $\Pi$:
$w=w^{(\sigma,S)}_{[a_1;\ldots;a_h]}$ and $w'=w^{(\sigma,S)}_{[a_1;\ldots;a_h;a]}$
\end{itemize}
Intuitively, $T$ is the set of transitions between states. A transition labelled $a$ from $w$ to $w'$ (represented by the triple
 $(w,a,w')$) is present in $T$ if, there is a (total) answer set of $\Pi$, in which $w$ is a state and the execution of action $a$ in $w$ leads to the state $w'$.

\section{Reasoning tasks}

Given a domain description $D=(\Pi,\cal C)$ over $\Sigma$ and a temporal {\em goal } $\alpha$ (a DLTL formula), we are interested
in finding out the extensions of  $D=(\Pi,\cal C)$ satisfying/falsifying $\alpha$.
While in the next section we will focus on the use of bounded model checking techniques for answering this question,
in this one we show that many reasoning problems, including temporal projection, planning
and diagnosis
can be characterized in this way.

Suppose that in Example \ref{es_shooting} we want to know if there is a scenario in which the turkey is not alive after the action sequence
$\neg  in\_sight?, wait; in\_sight?, load, shoot $.
We can solve this {\em temporal projection problem} by finding out an extension of the domain description which satisfies the temporal formula
$$\la \neg  in\_sight?; wait; in\_sight?; load; shoot \ra \neg alive$$

The extension $S_1$ in Example \ref{es_extension} indeed satisfies the temporal formula above, since
$\la \neg  in\_sight?; wait; in\_sight?; load; shoot \ra \neg alive$ is true in the linear model
$M_{S_1}=(\sigma_1,V_S)$ associated with the extension $S_1$.

It is well known that a planning problem can be formulated as a satisfiability problem \cite{Giunchiglia&Traverso99}.
In case of complete state and deterministic actions,
 the problem of finding a plan which makes the turkey not alive and the gun loaded,
can be stated as the problem of finding out an extension of the domain description in which the formula
$\Diamond ( \neg alive \wedge loaded)$
is satisfied.
The extension provides a plan for achieving the goal $ \neg alive \wedge loaded$.

With an incomplete initial state, or with nondeterministic actions,
the problem of finding a conformant/universal plan which works for all the possible completions of the initial state
and for all the possible outcomes of nondeterministic actions cannot be simply solved by checking the satisfiability of the formula above.
The computed plan must also be tested to be a conformant plan.
On the one hand, it must be verified that the computed plan $\pi$ always achieves the given $Goal$,
i.e., there is no extension of the domain description satisfying the formula $\la \pi \ra \neg Goal$.
On the other hand, it must be verified that $\pi$ is executable in all initial states.
This can be done, for instance, adopting techniques similar to those in \cite{Giunchiglia:KR2000}.
\cite{Leone03} addresses the problem of conformant planning in the DLV$^{\cal K}$ system.
\cite{Morales08} develops conformant planners based on a notion of approximation of action theories
in the action language ${\cal AL}$ \cite{Baral2000}.

As concerns diagnosis, consider systems like the one in Example \ref{es_sistema}.
A diagnosis of a fault observation $obs_f$ is a run from the initial state to a state in which $obs_f$ holds and which does not contain fault observations in the previous states \cite{Panati00}, i.e., an extension
satisfying the formula: $(\neg obs_1\wedge \ldots \wedge \neg obs_n)\;  \Unt  \; obs_f$,
where $obs_1$, $\ldots, obs_n$ are all the possible observations of fault.
In Example 3, $p\_obs\_low$ is the only possible fault observation, hence
a diagnosis for it is an extension of the domain description which
satisfies $\Diamond p\_obs\_low$.

As concerns property verification, an example has been given in Example 2.
We observe that the verification that a domain description $D$ is well-defined can be  done by
adding to the domain description  a static law
$\Box ( undefined\_fluent \leftarrow not ~ f \wedge not ~ \neg f)$, for each fluent literal $f$,
and by verifying that there are no extensions in which $\Diamond  undefined\_fluent$ holds
in the initial state.

Other reasoning tasks which can be addressed by checking the satisfiability or validity of formulas
in a temporal action theory are
multiagent protocol verification  \cite{GMS-JAL06},
and verification of the compliance of business processes to norms \cite{CLIMA2010}.

\section{Model checking and bounded model checking}

LTL is widely used to prove properties of
systems by means of {\em model checking}.
The property to be verified can be
represented as an LTL formula $\varphi$, whereas a Kripke structure provides the model of the system
to be verified (in the current case, the transition system associated with the domain description). A standard approach to verification is based on the construction of the B\"{u}chi automaton
for the negated property and on the computation of the product of such automaton with the model of the system.
The property is verified when the language accepted by the product automaton is empty,
whereas any infinite word accepted by the product automaton provides a counterexample to the validity of
$\varphi$. This approach is also feasible for DLTL, as it is possible to construct a B\"{u}chi automaton for a given DLTL formula \cite{Henriksen99}. In particular, as for LTL, the construction of the
automaton can be done on-the-fly, while checking for the emptiness of
the language accepted by the automaton \cite{AMAI06}.

In \cite{Cimatti03} it has been shown that, in some cases, model checking can be more
efficient if, instead of building the product automaton and checking for an
accepting run on it, we build only an accepting run of the automaton (if there is one).
This technique is called \emph{bounded model checking} (BMC), since it looks for
paths whose length is bounded by some integer $k$, by iteratively increasing
the length $k$ until a run satisfying $ \neg \varphi$ is found (if one exists).
The paths considered are infinite paths which can be finitely represented as
paths of length $k$ with a back loop from state $k$ to a previous state in the path: it can be shown that, if a  B\"{u}chi automaton has an accepting run,
it has one which can be represented in this way.

A BMC problem can be efficiently reduced to a propositional
satisfiability problem \cite{Cimatti03} or to an ASP problem \cite{Niemela03}.
BMC provides a partial decision procedure for checking validity:
if no model exists, the iterative procedure will never stop.
Techniques for achieving completeness are described in \cite{Cimatti03}.

In the next section, we address the problem of defining a translation of a domain description into standard  ASP,
so that bounded model checking techniques can be used to check if a temporal goal (a DLTL formula)  is satisfiable in some  extension of the domain description.
The approach we propose for the verification
of  DLTL formulas extends
the one developed in \cite{Niemela03} for bounded LTL model checking with Stable Models.

\section{Translation to ASP}

\label{sec:translation}

In this section, we show how to translate a domain description to standard ASP.

A temporal model consists of an infinite sequence of actions and a valuation function
giving the value of fluents in the states of the model.
States are represented in ASP as integers, starting with the initial state 0.
We will use the predicates
$occurs(Action,State)$ and $holds(Literal,State)$.
Occurrence of exactly one action in each state must be encoded:

$\neg occurs(A,S) \leftarrow occurs(A1,S),action(A),action(A1),A\neq A1,state(S). $

\indent
$occurs(A,S) \leftarrow not\; \neg occurs(A,S),action(A),state(S). $

Given a  \emph{domain description} $(\Pi,  {\cal C})$, the rules in $\Pi$
can be translated as follows.

\noindent
{Action laws}
$\Box ( [a] (\neg)f_0 \leftarrow t_1, \ldots ,t_m , not\; t_{m+1} , \ldots ,  not\; t_n) $
are translated to
$$(\neg)holds(f_0,S') \leftarrow state(S), S'=S+1,
occurs(a,S), h_1 \ldots h_m, not \; h_{m+1} \ldots not \; h_n $$
where $h_i = (\neg)holds((f_i,S')$ if $t_i = [a](\neg)f_i$ and $h_i = (\neg)holds(f_i,S)$ if $t_i =(\neg) f_i$.

\noindent
{Dynamic causal laws}
$\Box ( \bigcirc (\neg)f_0 \leftarrow t_1, \ldots ,t_m , not\; t_{m+1} , \ldots , not\; t_n) $
are translated to
$$(\neg) holds(f_0,S')  \leftarrow state(S), S'=S+1, h_1 \ldots h_m, not \; h_{m+1} \ldots not \; h_n $$
where
 $h_i = (\neg) holds( f_i,S')$ if $t_i = \bigcirc (\neg) f_i$ and $h_i =(\neg) holds( f_i,S)$ if $t_i =(\neg) f_i$.

\emph{Static causal laws} (\ref{sc_law})
are translated in a similar way (replacing $S'$ with $S$ in the head),
while \emph{initial state laws} are evaluated in state 0.

\noindent
{Precondition laws}
$\Box ( [a]\bottom \leftarrow l_{1}, \ldots ,l_m , not\; l_{m+1} , \ldots , not\; l_n) $
are translated to ASP constraints
$$ \leftarrow state(S), occurs(a,S),
   h_{1} \ldots h_m, not \; h_{m+1} \ldots not \; h_n $$
\noindent
where  $h_i = holds(l_i,S)$.

As described in the previous section, we are interested in infinite models represented
as \emph{k-loops}, i.e., finite sequences of states from 0 to $k$ with a back loop
from state $k$ to a previous state. Thus, we assume a bound $k$ on the
number of states.

The above rules compute a finite model from state 0 to state $k+1$.
To detect the loop, we must find a state $j$, $0 \leq j \leq k$, equal to state $k+1$
This can be achieved by defining a predicate
$eq\_last(S)$ to check if state $S$ ie equal to state $k+1$,
and a predicate $next(S1,S2)$ such that $next(i,i+1)$ for $0 \leq i \leq k-1$, and $next(k,j)$.

$\emph{diff}\_last(S) \leftarrow state(S), S<=k, fluent(F),
		holds(F,S), \neg holds(F,k+1). $
		
$\emph{diff}\_last(S):- state(S), S<=k, fluent(F),
		holds(F,k+1), \neg holds(F,S).$
		
$eq\_last(S):- state(S),  S<=k, not \; \emph{diff}\_last(S).$

$next(S,SN) \leftarrow  state(S),S<k, SN=S+1.$

$\neg next(k,S) \leftarrow  next(k,SS),state(S),state(SS),S \neq SS.$

$next(k,S) \leftarrow  state(S), S<=k, not \; \neg next(k,S).$

$\leftarrow next(k,S), not \; eq\_last(S).$

\noindent
The second and third rule for $next$ impose that is
exactly one state next to state $k$; the last constraint imposes that
such a state is equal to state $k+1$.

Given a domain description $(\Pi,  {\cal C})$, we denote by $tr(\Pi)$ the set of rules containing the translation
of each law in $\Pi$ as well as the definitions of $eq$,$\emph{diff}$ and $next$, as defined above.
Observe that an  answer set $R$ of $tr(\Pi)$ such that, for each state $i=1,\ldots,k$, either
${holds(p,i)} \in R$ or ${\neg holds(p,i)} \in R$,
represents a temporal model as a {\em k-loop}.
The temporal model, $M_R =(\sigma_R,V_R)$ associated with $R$ can be defined as follows:

$\sigma_R = a_1 a_2 \ldots a_j a_{j+1} \ldots a_{k+1} a_{j+1} \ldots a_{k+1} \ldots$

\noindent
where
$occurs(a_1,0),occurs(a_2,1),...,occurs(a_{j+1},j),...,occurs(a_{k+1},k),$ $next(k,j)$
(i.e., $a_{k+1}$ leads back to state $j$)
belong to $R$, and, for all atomic propositions $p\in {\cal P}$:

$p \in V_R(\varepsilon)$ if and only if  ${holds(p,0)} \in R$

 $p \in V_R(a_1 \ldots a_h)$ if and only if  ${holds(p,h)} \in R$, for $0 < h\leq k$

 $p \in V_R(a_1 \ldots a_{k+1})$ if and only if ${holds(p,j)} \in R$.

We can show that there is a one to one correspondence between the temporal answer sets of $\Pi$ and
the answer sets of the translation $tr(\Pi)$.
Let $(\Pi,  {\cal C})$ be a well-defined domain description over $\Sigma$.
\begin{theorem}\label{th1}
\begin{itemize}
\item[(1)] Given a temporal answer set $(\sigma,S)$ of $\Pi$ such that $\sigma$
can be finitely represented as a finite
path with a k-loop, there is a
consistent answer set $R$ of $tr(\Pi)$ such that
$R$ and $S$ correspond to the same temporal model.
\item[(2)] Given a consistent answer set $R$ of $tr(\Pi)$, there is a temporal answer set $(\sigma,S)$ of $\Pi$
(that can be finitely represented as a finite path with a back loop) such that
$R$ and $S$ correspond to the same temporal model.
\end{itemize}
We refer to Appendix A for the proof.
\end{theorem}

Let us now come to the problem of evaluating a DLTL formula over the models associated with the answer sets of
$tr(\Pi)$.
To deal with DLTL formulas,
we use the predicate $sat(alpha,S)$,
to express satisfiability of a DLTL formula $\alpha$ in a state of a model.
As in \cite{AMAI06} we assume that $until$ formulas are indexed with finite automata rather than regular expressions,
by exploiting the equivalence between
regular expressions and finite automata.
Thus, we have $\alpha {\cal U}^{{\cal A}(q)} \beta$ instead of $\alpha
{\cal U}^\pi \beta$, where ${\cal L}({\cal A}(q)) = [[\pi]]$.
More precisely, let ${\cal A}=(Q,\delta,Q_F)$ be
an $\epsilon$-free nondeterministic finite automaton over the
alphabet $\Sigma$ without an initial state, where $Q$ is a finite
set of states, $\delta: Q \times \Sigma \rightarrow 2^Q$ is the
transition function, and $Q_F$ is the set of final states. Given
a state $q \in Q$, we denote with ${\cal A}(q)$ an automaton
${\cal A}$ with initial state $q$.
In the definition of predicate $sat$ for \emph{until} formulas, we refer to the
following axioms \cite{Henriksen99}:
\begin{itemize}
\item[] $\alpha {\cal U}^{{\cal A}(q)} \beta \equiv (\beta \vee
(\alpha \wedge \bigvee_{a\in\Sigma}\langle a \rangle
\bigvee_{q'\in \delta(q,a)} \alpha {\cal U}^{{\cal A}(q')} \beta
))$ \ \ \ ($q$ is a final state of ${\cal A}$)

\item[] $\alpha {\cal U}^{{\cal A}(q)} \beta \equiv ( \alpha \wedge
\bigvee_{a\in\Sigma}\langle a \rangle \bigvee_{q'\in \delta(q,a)}
\alpha {\cal U}^{{\cal A}(q')} \beta) $ \ \ \ \ \ ($q$ is not a final state of ${\cal A}$)
\end{itemize}

In the translation to ASP, DLTL formulas will be represented with terms.
In particular, the formula $\alpha {\cal U}^{{\cal A}(q)} \beta$ will be represented as  $until(A,q,alpha,beta)$.
Furthermore, we assume the automaton ${\cal A}$ to be described with the predicates
$trans(A,Q1,$ $Act,Q2)$ defining the transitions, and $final(A,Q)$ defining the
final states.
The definition of $sat$ is the following:
\vspace{-0.1cm}
\begin{tabbing}
TuntilNFSSS \=  XXX \= \kill

fluent: \> $sat(F,S) \leftarrow fluent(F), holds(F,S).$ \\
or: \> $sat(or(Alpha, Beta),S) \leftarrow sat(Alpha,S).$ \\
\> $sat(or(Alpha, Beta),S) \leftarrow sat(Beta,S).$\\
neg: \> $sat(neg(Alpha),S) \leftarrow not \;sat(Alpha,S).$ \\
until: \> $sat(until(Aut,Q,Alpha,Beta),S) \leftarrow final(Aut,Q),sat(Beta,S).$\\
 \> $sat(until(Aut,Q,Alpha,Beta),S) \leftarrow $\\
\> \> $sat(Alpha,S),trans(Aut,Q,Act,Q1),occurs(Act,S),$\\
\> \> $next(S,S1),sat(until( Aut,Q1,Alpha,Beta),S1).$

\end{tabbing}
Similar definitions can be given for derived connectives and modalities.
For instance, the temporal formulas $\Diamond \alpha$, $\la a\ra \alpha$ and $[a]\alpha$ are represented,
respectively, by the terms
$eventually(t\_alpha)$, $diamond(a,t\_alpha)$ and $box(a,t\_alpha)$,
where $t\_alpha$ is the term encoding the formula $\alpha$.
The definition of  $sat$ for such formulas is the following:
\begin{tabbing}
TuntilNFSS \=  XXX \= \kill

eventually: \> $sat(eventually(Alpha),S) \leftarrow sat(Alpha,S).$\\
eventually: \> $sat(eventually(Alpha),S) \leftarrow next(S,S1),sat(eventually(Alpha),S1).$\\

$\la a\ra$: \> $sat(diamond(A,Alpha),S) \leftarrow occurs(A,S),next(S,S1),sat(Alpha,S1).$\\

$[a]$: \> $sat(box(A,Alpha),S) \leftarrow action(A), occurs(B,S),A!=B.$\\

$[a]$: \> $sat(box(A,Alpha),S) \leftarrow occurs(A,S),next(S,S1),sat(Alpha,S1).$

\end{tabbing}
Since states are complete, we can identify default negation with classical negation,
thus
having a two valued interpretation of DLTL formulas.
We must also add a constraint
$\leftarrow not \; sat(t\_alpha,0)$,
for each temporal
constraint $\alpha$ in the domain description, where states are represented by numbers,
$0$ is the initial state and $t\_alpha$ is the term encoding the formula $\alpha$.
The presence of the constraint $\leftarrow\;not \; sat(t\_alpha,0)$,
in the translation of the domain description guarantees that $\alpha$ must be satisfied,
as the negated formula  $not \; sat(t\_alpha,0)$ is not allowed to be true in the answer set.

As an example, the encoding of the temporal constraint
$$\Box [begin] \la sense(a); sense(b); (deliver(a)+ deliver(b) + wait ); begin \ra \top $$
in Example \ref{es_mailbox}, is given by the following rules:

$\leftarrow not \; sat(neg(ev(neg(box(begin,until(aut,q1,true,true))))),0).$

$trans(aut,q1,sense(a),q2).$

$trans(aut,q2,sense(b),q3).$

$trans(aut,q3,deliver(a),q4).$

$trans(aut,q3,deliver(b),q4).$

$trans(aut,q3,wait,q4).$

$trans(aut,q4,begin,q5).$

$final(aut,q5).$

\noindent
The first rule encodes the constraint, while the following ones encode the definition of the automaton
$aut$, which is equivalent to the regular expression indexing the until formula in the constraint.

It is easy to see that the computation of the satisfiability of a formula $\alpha$ in a
given state depends only on a finite set of formulas consisting of the subformulas of $\alpha$ and the formulas
derived from an \emph{until} subformula.
We say that a  formula $\gamma {\cal U}^{{\cal A}(q')} \beta$ is
\emph{derived} from a formula $\gamma {\cal U}^{{\cal A}(q)} \beta$ if $q'$ is
reachable from $q$ in $\cal A$.

It is possible to see that the definition of the predicate $sat$, as given above for the base cases (fluent, or, neg, until),
provides a correct evaluation of the temporal formulas
over the temporal models associated with the translation $tr(\Pi)$ of $\Pi$.
Let $tr'(\Pi)$ be the set of rules extending the rules in $tr(\Pi)$ with the definition of predicate   $sat$ above.
Let $(\Pi,  {\cal C})$ be a well-defined domain description over $\Sigma$.
We can prove the following theorem (the proof can be found in Appendix A):
\begin{theorem} \label{SAT}
Let $\Pi$ be the set of laws of a well-defined domain description,
$R$ an answer set of $tr(\Pi)$ and $\alpha$ a DLTL formula.
The temporal model $M_R=(\sigma,V)$ associated with $R$ satisfies $\alpha$ if and only if there is an answer set $R'$ of $tr'(\Pi)$
such that $R \subset R'$ and $sat(t\_alpha,0)\in R'$
(where  $t\_alpha$ is the term representing the formula $\alpha$
and $trans$ and $final$ encode the automata indexing the until formulas in $\alpha$).
\end{theorem}

The above formulation of $sat$ is indeed the direct translation of the semantics of DLTL, which is given for infinite models.
Intuitively, we can show that it works also when the model is represented as a {\em k-loop}, by considering the case of {\em until} formulas.
If $S$ is a state belonging to the loop, the goal  $sat(\alpha {\cal U}^{{\cal A}(q)}
\beta,S)$ can depend cyclically on itself. This happens if the only rule which can be
applied to prove the satisfiability of  $\alpha {\cal U}^{{\cal A}(q)} \beta$ (or one of its
derived formulas in each state of the loop) is the second rule of \emph{until}.
In this case, $sat(\alpha {\cal U}^{{\cal A}(q)} \beta,S)$ will be undefined, which
amounts to say that $\alpha {\cal U}^{{\cal A}(q)} \beta$ is not true.
This is correct, since, if this happens, $\alpha$ must be true in each state of the loop,
and $\beta$ must be false in all states of the loop corresponding to final states of $
{\cal A}$. Thus, by unfolding the cyclic sequence into an infinite sequence, $\alpha
{\cal U}^{{\cal A}(q)} \beta$ will never be satisfied.

Given a domain description $D=(\Pi,{\cal C})$, the translation $tr(D)$ of $D$ contains:
the translation $tr(\Pi)$ of $\Pi$, the definition of the predicates $sat$, $trans$ and $final$,
and, for each temporal formula
$\alpha$ in ${\cal C}$, the constraint $\leftarrow not \; sat(t\_alpha,0)$.

Let $(\Pi, {\cal C})$ be a well-defined domain description over $\Sigma$.
Given Theorems \ref{th1} and \ref{SAT} above, it can be proved that:
\begin{corollary}
There is a one to one correspondence between the extensions of the domain
description $D$ and
the answer sets of its translation $tr(D)$ in ASP.
\end{corollary}
More precisely, each extension of $D$ is in a one to one correspondence with an answer set of $tr(D)$, and both of them are
associated with the same temporal model.

Given a temporal formula $\alpha$, we may want to check if there is an extension of the domain description $D$
satisfying it. To this purpose, as for the temporal formulas in ${\cal C}$,
we add to the translation $tr(D)$ of $D$ the constraint
$\leftarrow not \; sat(t\_alpha,0)$, so that the answer sets falsifying $\alpha$ are excluded.

According to the bounded model checking technique,
the search for an extension of the domain description satisfying $\alpha$ is done
by iteratively increasing the
length $k$ of the sequence searched for, until a cyclic model is found (if one exists).
On the other hand, validity of a formula $\alpha$ can be proved, as usual in model
checking, by verifying that $D$ extended with $\neg \alpha$ is not satisfiable.
Let us consider, from Example \ref{es_mailbox}, the property $\Box (mail(b) \supset \Diamond \neg mail(b))$
(if there is mail for $b$, the agent will eventually deliver it to $b$).
This formula is valid if its negation $\Diamond \neg (mail(b) \supset \Diamond \neg mail(b))$
is not satisfiable. We verify the satisfiability of this formula,
by adding to the translation of the domain description the constraint

 $\leftarrow not \; sat(ev(neg(impl(mail(b),ev(neg(mail(b)))))),0).$

 \noindent
 and looking for an extension.
 The resulting set of rules indeed has extensions,
 which can be found for $k\geq 3$ and provide counterexamples
 to the validity of the property above. For instance, the extension in which
 $next(0,1), next(1,2), next(2,3)$, $next(3,0),$ $occurs(begin,$ $0)$, $occurs(sense\_mail(a),$ $1)$,
 $occurs(sense\_mail(b),$ $2)$,
 $occurs(deliver\_mail(a),3)$, where $mail(b)$ holds in all states,
 and \texttt{mail(a)} only in states 2 and 3, can be obtained for $k=3$.

In Appendix B we provide the encoding of BMC and of Example \ref{es_mailbox} in the DLV-Complex extension
(https://www.mat.unical.it/dlv-complex) of DLV \cite{LeoneDLV}.
In Appendix C we report tests of our approach for bounded model checking of
DLTL formulas in the line of the LTL BMC experiments in \cite{Niemela03}.
Results are provided for a DLV encoding of BMC and of action domain descriptions for the dining philosophers problems considered in that paper.
The scalability of the two approaches is similar.

\section{Conclusions and related work}

In this paper we have described an action language which is based on a temporal extension of ASP,
in which temporal modalities are included within rules. In the action language,
general temporal DLTL formulas (possibly including regular programs indexing temporal modalities)
are allowed in the domain description to constrain the space of possible extensions.
The approach naturally deals with non-terminating computations and relies on bounded
model checking techniques for the verification of temporal formulas.
In \cite{GMS00} a temporal action theory  has been developed, which is based on the linear temporal logic DLTL
and adopts a monotonic solution to the frame problem based on completion.
Due to the different treatment of the frame problem, even in the case when default negation
is not present in the body of the  laws in $\Pi$,
the notion of extension defined here is not equivalent to the one in \cite{GMS00}, which
requires action and causal laws to be stratified to avoid
unexpected extensions due to cyclic dependencies.

Bounded model checking (BMC) \cite{Cimatti03} is based on the idea to search for a counterexample of the property to be checked
in executions which are bounded by some integer $k$.
SAT-based BMC methods do not suffer from the state
explosion problem as the methods based on BDDs.
\cite{Niemela03}
exploit BMC in the verification of
asynchronous systems modeled by 1-safe Petri nets.
They provide a translation of a Petri net to a logic program which captures
the execution of the net up to $n$ steps and
they develop a compact encoding of BMC of LTL formulas
as the problem of finding stable models of logic programs.
As a difference, the work in this paper aims at verifying properties of a temporal action theory
including DLTL temporal constraints.
Hence, we provide a translation of the action theory into ASP
and
we extend the encoding of BMC in
\cite{Niemela03}
to deal with DLTL formulas.

Our encoding of BMC of LTL formulas in ASP does not make use of the B\"{u}chi automaton construction
 to build the path satisfying a formula. As future work, we aim at exploring an alternative approach
which exploits  the B\"{u}chi automaton of the formula to achieve completeness of BMC.

Stemming from the seminal paper of Gelfond and Lifschitz on the action language ${\cal A}$
\cite{Gelfond&Lifschitz:93},
a lot of work has been devoted to define logic-based action languages.
In particular, ASP has been shown to
be well suited for reasoning about dynamic domains
\cite{Gelfond}. \cite{Baral2000} provide an encoding in ASP of the action
specification language  ${\cal AL}$, which extends the action description language ${\cal A}$
by allowing static and dynamic causal laws, executability conditions and
concurrent actions.
The proposed approach has
been used for planning \cite{Morales08} and diagnosis \cite{Balduccini03}.

The action language defined in this paper
can be regarded as a temporal extension of the language
${\cal A}$ \cite{Gelfond&Lifschitz:93} which allows for
general temporal constraints, complex actions and infinite computations, but
does not deal with concurrent actions  nor with incomplete knowledge.
As regards laws in $\Pi$,
our temporal action language has strong relations with the action languages ${\cal K}$ and ${\cal C}$.

The logic-based planning language, ${\cal K}$ \cite{Leone03,Leone04} is well suited for planning under incomplete knowledge and allows for concurrent actions.
The main construct of ${\cal K}$ are causation rules of the form:
\texttt{caused  f  if  B after A},
meaning ``If \texttt{B} is known to be true in the current state and \texttt{A} is known to be true in the previous state, then \texttt{f} is known to be true in the current state''.
Default  negation can be used in the body of the rules and \texttt{A} may contain action atoms.
The semantics of planning domains is defined in terms of states and transitions.

The temporal action language introduced in Section 3 for defining the component $\Pi$ of the domain description
can be regarded, apart from minor differences, as a fragment of ${\cal K}$ in which concurrent actions are not allowed.
In particular, action laws (\ref{a_law}) and dynamic causal laws (\ref{dc_law}):
\begin{quote}
$\Box([a]l_0 \leftarrow (not) l_1,\ldots, (not) l_m, (not)[a]l_{m+1}, \ldots, (not)[a]l_k)$

$\Box(\bigcirc l_0 \leftarrow (not) l_1,\ldots, (not) l_m, (not) \bigcirc l_{m+1}, \ldots, (not) \bigcirc l_k)$

\end{quote}
can be mapped to the causation rules:
\begin{quote}
\texttt{{\bf caused} $l_0$  {\bf if} $ (not) l_{m+1} \ldots, (not)l_k$  {\bf  after} $a,  (not) l_1,\ldots, (not) \;l_m$ }

\texttt{{\bf caused} $l_0$  {\bf if} $ (not) l_{m+1} \ldots, (not)l_k$  {\bf  after} $(not) l_1,\ldots, (not) \;l_m$ }

\end{quote}
with the proviso, for dynamic causal laws, that $m\geq 1$.
In case the literals $(not) l_1,\ldots,$ $ (not) l_m$ are not present
(and the \texttt{{\bf after}} part of the causation rule is empty),
our dynamic causal law does not produce any effect on the initial state (which
is not the next state of any other state) while the causation rule does.
For this reason,  our static causal laws
can then be mapped to causation rules with empty \texttt{{\bf after}}  part.
A similar translation can be given to precondition laws, which are special kinds of action laws
and to initial state laws, which can be mapped to initial state causation rules in ${\cal K}$.
All actions are regarded as being always executable,
i.e., \texttt{{\bf executable}  a}, holds for all actions \texttt{a}.
The correctness of this mapping emerges form the ASP encoding of our temporal language, which is similar,
apart from minor differences, to the translation of ${\cal K}$ to answer set programming \cite{Leone03}.

The system $DLV^{\cal K}$ provides an implementation of ${\cal K}$
on top of the disjunctive logic programming system DLV.
$DLV^{\cal K}$ does not only solve optimistic planning problems, but also secure planning problems under incomplete initial states (\emph{conformant planning}).
$DLV^{\cal K}$ does not appear to support other kinds of reasoning besides planning, and, in particular, does not allow to express or verify temporal properties nor to reason about infinite computations.

The languages ${\cal C}$ and ${\cal C}^+$ \cite{Giunchiglia&al:98,Giunchiglia&al:2004}
also deal with actions with indirect and nondeterministic effects
and with concurrent actions, and are based on nonmonotonic causation rules syntactically similar to those of ${\cal K}$, where head and body of causation rules can be boolean combinations of atoms.
Their semantics is based on a nonmonotonic causal logic
\cite{Giunchiglia&al:2004}. Causation rules can be represented in this logic by indexing fluents and actions with an integer $i$ ($i = 0, \ldots n$), in such a way that models of the causal theory correspond to histories of length $n$.
The semantics of causal logic requires states to be complete.
Due to the differences between the underlying semantics, a mapping between our action language
and the languages ${\cal C}$ and ${\cal C}^+$ appears not to be straightforward.

If a ${\cal C}^+$ causal theory is definite (the head of a rule is an atom), it is possible to reason about it by turning the theory into a set of propositional formulas by means of a completion process, and then invoke a satisfiability solver.
In this way it is possible to perform various kinds of reasoning such as prediction, postdiction or planning.
However, the language does not exploits standard temporal logic constructs to reason about actions.

In the context of planning, temporally extended goals allow the specification of properties that
have to be achieved in the states along the execution of the plan. The need for state trajectory constraints
has been advocated, for instance, in PDDL3  \cite{Gerevini05}, the domain description language used in the 2006
International Planning Competition. \cite{BacchusAIJ00} exploits
a first order linear temporal logic for defining domain dependent search control knowledge
in the planner TLPlan.
\cite{Pistore01}
define a planning algorithm that generates plans for extended goals in a nondeterministic
domain, where extended goals are CTL formulas.
\cite{Son06TCL} shows that temporal control knowledge can be incorporated in a planner written in ASP.
It provides a translation  of a planning problem whose domain is defined in the action language ${\cal B}$ into ASP as well as a translation of the temporal constraints on the domain.
The work on temporally extended goals
in \cite{Traverso02,Baral07} is concerned with expressing preferences among goals and
exceptions in goal specification.
\cite{Son06TPLP,Bienvenu06} introduce languages including temporal operators for expressing
preferences on solutions of planning problems.
\cite{Son06TPLP}, in particular, builds on answer set planning, i.e., computing plans in
ASP; the computation of {\em preferred}
plans is also mapped to ASP, relying on an optimization predicate.
As a difference with the proposals above, in this paper we do not specifically focus on planning.
Our language is intended to address
several different reasoning tasks (including property verification)
on rich domain descriptions, allowing for ramifications,
nondeterministic and complex actions, incomplete initial
states, and, in particular, it can be used for reasoning about infinite computations.
However, in this paper,  we do not address the problem of expressing preferences among goals.

As our language includes program expressions in the temporal formulas, it is related to the Golog language
\cite{Levesque97}, in which complex actions (plans) can be formalized as Algol-like programs.
$\cal{ESG}$ \cite{Classen:08} is a second order extension of CTL* for reasoning about
nonterminating Golog programs.
In $\cal{ESG}$ programs include, besides regular expressions, nondeterministic choice of arguments and concurrent
composition. The paper presents a method for verification of a first order CTL fragment of $\cal{ESG}$, using
model checking and regression based reasoning. Because of first order quantification, this fragment is
in general undecidable. DLTL \cite{Henriksen99} can be regarded as a decidable LTL fragment of $\cal{ESG}$.
Satisfiability in DLTL is known to be PSPACE-complete  \cite{Henriksen99}, as for LTL.

Observe that, although our temporal answer sets are, in general, infinite, we do not need to exploit
specific techniques for reasoning about infinite answer sets \cite{Bonatti04},
due to the property that an infinite path can be finitely represented as a k-loop.

\begin{center}
\noindent {\Large \bf Acknowledgments}
\end{center}

We thank the anonymous referees for their helpful comments. This work has been
partially supported by Regione Piemonte, Project ICT4LAW
(\emph{ICT Converging on Law: Next Generation Services for
Citizens, Enterprises, Public Administration and Policymakers}).

\newpage
\begin{appendix}

\begin{center}
\noindent {\Large \bf Appendix A}
\end{center}

\medskip

We prove Theorem 1 and Theorem 2.
Let $(\Pi,  {\cal C})$ be a well-defined domain description over $\Sigma$.
We show that there is a one to one correspondence between the temporal answer sets of $\Pi$ and
the answer sets of the translation $tr(\Pi)$.
\medskip

{\em Theorem \ref{th1}}
\begin{itemize}
\item[(1)] Given a temporal answer set $(\sigma,S)$ of $\Pi$ such that $\sigma$
can be finitely represented as a finite
path with a k-loop, there is a
consistent answer set $R$ of $tr(\Pi)$ such that
$R$ and $S$ correspond to the same temporal model.
\item[(2)] Given a consistent answer set $R$ of $tr(\Pi)$, there is a temporal answer set $(\sigma,S)$ of $\Pi$
(that can be finitely represented as a finite path with a back loop) such that
$R$ and $S$ correspond to the same temporal model.
\end{itemize}

\begin{proof}
Let us prove item (1).
Let $(\sigma,S)$ be a temporal answer set of $\Pi$ such that $\sigma$
can be finitely represented as a finite
path with a back loop, i.e.,
$$\sigma = a_1 a_2 \ldots a_j a_{j+1} \ldots a_{k+1} a_{j+1} \ldots a_{k+1} \ldots$$
We construct an answer set $R$ of $tr(\Pi)$ as follows.
$R$ contains the following literals:

$state(0), \ldots, state(k)$

$next(0,1), next(1,2), \ldots, next(k-1,k), next(k,j),$
$\neg next(k,s)$, for all $s \neq j$

$occurs(a_1,0),occurs(a_2,1),...,occurs(a_{j+1},j),...,occurs(a_{k+1},k)$

$\neg occurs(a,s)$, for all other ground instances of predicate {\em occurs},

$eq\_last(j)$

\noindent
for all $i = 0,\ldots,k$, for all fluent names $f \in {\cal P}$:
\begin{center}
$(\neg) holds(f,i) \in R$ if and only if  $[a_1;\ldots; a_i](\neg)f \in S$
\end{center}
From the consistency of $S$, it is easy to see that $R$ is a consistent set of literals.
To show that $R$ is an answer set of $tr(\Pi)$, we show that:
(i)  $R$ is closed under $tr(\Pi)^R$;
(ii) $R$ is minimal (in the sense of set inclusion) among the consistent sets of literals closed under $tr(\Pi)^R$.

(i) For all the rules $r$ in $tr(\Pi)^R$, we have to prove that if the literals in the body of $r$
belong to $R$, then the head of $r$ belongs to $R$.
Let us consider the case when the rule $r$ in $tr(\Pi)^R$ is obtained by translating an action law in $\Pi$,
of the form:
$$\Box ( [a] (\neg) f \leftarrow t_1, \ldots ,t_m , not\; t_{m+1} , \ldots ,  not\; t_n) $$
(the other cases are similar). In this case, $tr(\Pi)$ contains  the translation of the action law above:
$$(\neg) holds(f,S') \leftarrow state(S), S'=S+1,
occurs(a,S), h_1 \ldots h_m, not \; h_{m+1} \ldots not \; h_n $$
where $h_i = (\neg) holds(f_i,S')$ if $t_i = [a](\neg) f_i$ or $h_i = (\neg) holds(f_i,S)$ if $t_i =(\neg) f_i$.

Let us consider the ground instantiation of the rule above from which $r$ is obtained.
Suppose $S$ is instantiated with some $s\in \{1,\ldots,k\}$.
It must be the case that $a=a_{s+1}$, as $occurs(a_{s+1},s)\in R$ and no other action occurs in state $s$ according to $R$.

If the rule $r$:
\begin{equation}
(\neg) holds(f,s+1) \leftarrow state(s), occurs(a_{s+1},s), h'_1 \ldots h'_m    \label{trad_action_law}
\end{equation}
belongs to the reduct $tr(\Pi)^R$ (where each $h'_t$ is the ground instantiation of $h_t$ with $S=s$),
then
 $h'_{m+1} \not \in R, \; \ldots,  \; h'_n \not \in R$.

We have to show that, if the body of  (\ref{trad_action_law}) belongs to $R$ then its head also belongs to $R$.

Assume $h'_1, \ldots, h'_m$ belong to $R$.
For each $i=1\ldots, m$, either $h'_i=  (\neg) holds(f_i,s)$ (if $t_i =(\neg) f_i$) or $h'_i = (\neg) holds(f_i,s+1)$ (if $t_i = [a](\neg) f_i$).
If $h'_i=  (\neg) holds(f_i,s)$, by construction of $R$, $[a_1;\ldots; a_s](\neg) f_i \in S$, i.e.,
$(\sigma, S), a_1, \ldots, a_s \models (\neg) f_i$, and hence, $(\sigma, S), a_1, \ldots, a_s \models t_i$.
If $h'_i =(\neg) holds(f_i,s+1)$,  by construction of $R$, $[a_1;\ldots; a_s; a_{s+1}](\neg) f_i \in S$, i.e.,
$(\sigma, S), a_1, \ldots, a_s, a_{s+1} \models (\neg) f_i$, hence, $[a_1;\ldots; a_s;$ $a_{s+1}](\neg) f_i \in S$,
and then $(\sigma, S), a_1, \ldots, a_s  \models [ a_{s+1}] (\neg) f_i$. Thus, $(\sigma, S), a_1, \ldots, a_s \models t_i$.
So the positive literals in the temporal action law are satisfied.

To show that the negated literals $ t_{m+1} , \ldots ,  t_n $ in the body of the temporal clause
are not satisfied in $(\sigma,S)$ at $a_1, \ldots, a_s$, consider the fact that
$h'_{m+1} \not \in R, \; \ldots,  \; h'_n \not \in R$.
Again, for each $i=m+1,\ldots, k$, either $h'_i=  (\neg) holds(f_i,s))$
or $h'_i = (\neg) holds(f_i,s+1)$.

If $h'_i=  (\neg) holds(f_i,s)\not \in R$, by construction of $R$, $[a_1;\ldots; a_s](\neg)f_i \not \in S$, i.e.,
$(\sigma, S), a_1, \ldots, a_s\not \models (\neg)f_i$, and hence, $(\sigma, S), a_1, \ldots, a_s \not \models t_i$.
If $h'_i = (\neg) holds(f_i,s+1)\not \in R$,  by construction of $R$, $[a_1;\ldots; a_s; a_{s+1}](\neg)f_i \not \in S$, hence,
$(\sigma, S), a_1, \ldots, a_s  \not \models [ a_{s+1}] (\neg)f_i$. Thus, $(\sigma, S), a_1, \ldots, a_s \not \models t_i$.

We have shown that the body of the temporal rule
$$\Box ( [a] (\neg)f \leftarrow t_1, \ldots ,t_m , not\; t_{m+1} , \ldots ,  not\; t_n) $$
is true in $(\sigma,S)$ at $a_1, \ldots, a_s$, i.e.,
 $$(\sigma, S), a_1, \ldots, a_s \models  t_1, \ldots ,t_m , not\; t_{m+1} , \ldots ,  not\; t_n$$
 As the temporal rule belongs to $\Pi$ and is satisfied in $(\sigma, S)$, we can conclude that its head is
 also satisfied in $a_1, \ldots, a_s$, i.e., $(\sigma, S), a_1, \ldots, a_s \models [a] (\neg )f $,
 namely, $[a_1; \ldots; a_s; a]  (\neg )f  \in S$.
 As we observed above, $a=a_{s+1}$, hence, $[a_1; \ldots; a_s; a_{s+1}]  (\neg )f \in S $ and,
 by construction of $R$, $(\neg)holds( f , s+1) \in R$.

To prove (ii), we have to show that $R$ is minimal (in the sense of set inclusion) among the consistent sets of literals closed under $tr(\Pi)^R$. Suppose $R$ is not minimal, and there is a consistent set of literals $R'$ which is closed under  $tr(\Pi)^R$ and such that $R' \subset R$.

Suppose there is a literal $A\in R$ such that $A \not \in R'$.
For the auxiliary predicates $occurs$, $next$, etc., it is easy to see that this cannot be the case.
Let us consider the case $A=(\neg)holds(f,i)$ and suppose that $(\neg)holds(f,i)\in R$ and $(\neg)holds(f,i) \not \in R'$.

We show that we can construct from $R'$ an $S'\subset S$ such that $(\sigma,S')$ satisfies the rules in $\Pi^{(\sigma,S)}$.
We define $S'$ as follows:
\begin{center}
$[a_1;\ldots; a_h](\neg)f \in S'$
if and only if
$(\neg) holds(f,h) \in R'$
\end{center}
It can be shown that $(\sigma,S')$ satisfies the rules in $\Pi^{(\sigma,S)}$.
In fact, for each rule $r$ in  $\Pi^{(\sigma,S)}$ whose body is satisfied in $(\sigma,S')$, there is a rule $r'$ in
$tr(\Pi)^R$, whose body is true in $R'$.
As $R'$ is closed under $tr(\Pi)^R$, the head of $r'$ must be true in $R'$.
By construction of $S'$, the head of $r$ is satisfied in $(\sigma,S')$.

As $S' \subset S$ and $(\sigma,S')$ satisfies the rules in $\Pi^{(\sigma,S)}$,
$S$ is not minimal among the interpretations $S''$ such that $(\sigma,S'')$ satisfies
 the rules in $\Pi^{(\sigma,S)}$.
This contradicts the hypothesis that $(\sigma,S)$ is a temporal answer set of $\Pi$.

\medskip
As the domain description is well-defined, $(\sigma, S)$ has to be a total temporal answer set.
Hence, for each state $i=1,\ldots,k$, either
${holds(p,i)} \in R$ or ${\neg holds(p,i)} \in R$.
It is easy to see that $R$ and $(\sigma, S)$ correspond to the same temporal model, as $M_S$ and $M_R$
are defined over the same sequence $\sigma$ and, for each finite prefix $\tau$ of $\sigma$, they give the same evaluation to atomic propositions in $\tau$.

\medskip
\medskip

Let us prove item (2).

Let $R$ be an answer set of $tr(\Pi)$.
We define a temporal answer set $(\sigma,S)$ of $\Pi$ as follows.

Given the definition of the predicates $next$ and $occurs$ in $tr(\Pi)$,
$R$ must contain, for some $k$ and $j$, and for some $a_1, \ldots , a_{k+1}$, the literals:

$next(0,1), next(1,2), \ldots, $ $next(k-1,k), next(k,j),$

$occurs(a_1,0),occurs(a_2,1),...,occurs(a_{j+1},j),..., $
$occurs(a_{k+1},k)$,

$eq\_last(j)$.

\noindent
We define $\sigma$ as:
$$\sigma = a_1 a_2 \ldots a_j a_{j+1} \ldots a_{k+1} a_{j+1} \ldots a_{k+1} \ldots$$
We determine the temporal literals that belong to $S$ as follows:
for all $i = 0,\ldots,k$ for all fluent names $f \in {\cal P}$:
\begin{center}
$[a_1;\ldots; a_i](\neg)f \in S$ if and only if $(\neg) holds(f,i) \in R$
\end{center}

From the consistency of $R$, it is easy to see that $S$ is a consistent set of temporal literals.
To show that $S$ is a temporal answer set of $\Pi$, we show that:\\
(i)  $(\sigma,S)$ satisfies all the rules in $\Pi^{(\sigma,S)}$;\\
(ii) $S$ is minimal (in the sense of set inclusion)
among the $S'$ such that $(\sigma,S')$ is a partial interpretation
satisfying the rules in  $\Pi^{(\sigma,S)}$.

\medskip

(i) Let us prove that  $(\sigma,S)$ satisfies all the rules in $\Pi^{(\sigma,S)}$.
Let
$$[a_1,\ldots, a_s] ( H \leftarrow t_1, \ldots ,t_m) $$
be a rule in $\Pi^{(\sigma,S)}$, where $a_1,\ldots a_s \in \mbox{\em prf}(\sigma)$.
Then there must be a law in $\Pi$ of the form:
$$\Box ( H \leftarrow t_1, \ldots ,t_m , not\; t_{m+1} , \ldots ,  not\; t_n) $$
such that $(\sigma,S), a_1\ldots a_s\not  \models t_i$, for $i={m+1},\ldots,n$.

Let us consider the case where such a law is a dynamic causal law,
(the other cases are similar). In this case,  $H=  \bigcirc (\neg) f$ and the law has the form:
$$\Box ( \bigcirc (\neg) f \leftarrow t_1, \ldots ,t_m , not\; t_{m+1} , \ldots ,  not\; t_n) $$
where, for all $i=1,\ldots, n$, $t_i = (\neg) f_i$ or $t_i = \bigcirc (\neg) f_i$.

Then, $tr(\Pi)$ contains  its translation:
$$(\neg) holds(f,S') \leftarrow state(S), S'=S+1,
h_1 \ldots h_m, not \; h_{m+1} \ldots not \; h_n $$
where $h_i = (\neg) holds(f_i,S')$ (if $t_i = \bigcirc (\neg) f_i$) or $h_i = (\neg) holds(f_i,S)$ (if $t_i =(\neg) f_i$).

Let us consider the ground instantiation of the rule above with $S=s$, for some $s\in \{1,\ldots,k\}$.
$$(\neg) holds(f,s+1) \leftarrow state(s), h'_1 \ldots h'_m, not \; h'_{m+1} \ldots not \; h'_n $$
where $h'_i = (\neg) holds(f_i,s+1)$ (if $t_i = \bigcirc (\neg) f_i$) or $h'_i = (\neg) holds(f_i,s)$ (if $t_i =(\neg) f_i$).

The rule
\begin{equation}
(\neg) holds(f,s+1) \leftarrow state(s), h'_1 \ldots h'_m  \label{trad_causal_law}
\end{equation}
must belong to the reduct $tr(\Pi)^R$.
In fact, we can prove that  $h'_{m+1} \not \in R, \; \ldots,  \; h'_n \not \in R$.
Let $t_i =(\neg) f_i$ and $h'_i = (\neg) holds(f_i,s)$.
From the hypothesis, we know that
$(\sigma,S), a_1\ldots a_s \not \not  \models t_i$, i.e.,
$(\sigma,S), a_1\ldots a_s \not \not  \models (\neg) f_i $, i.e.,
$ [a_1;\ldots; a_s](\neg) f_i \not \in S$.
As, by construction of $(\sigma,S)$,
$ (\neg) holds(f_i,s) \in R$ iff $ [a_1;\ldots; a_s](\neg) f_i \in S$,
we conclude $ (\neg) holds(f_i,s) \not \in R$.
Let $t_i = \bigcirc (\neg) f_i$ and $h'_i = (\neg) holds(f_i,s+1)$.
From the hypothesis, we know that
$(\sigma,S), a_1\ldots a_s \not  \models t_i$, i.e.,
$(\sigma,S), a_1\ldots a_s \not  \models \bigcirc (\neg) f_i $, i.e.,
$ [a_1;\ldots; a_s; a_{s+1}](\neg) f_i \not \in S$.
As, by construction of $(\sigma,S)$,
$ (\neg) holds(f_i,s+1) \in R$ iff $ [a_1;\ldots; a_{s+1}](\neg) f_i \in S$,
we conclude $ (\neg) holds(f_i,s+1) \not \in R$, that is $h'_i \not \in R$.

To show that the law
$$[a_1,\ldots, a_s] ( H \leftarrow t_1, \ldots ,t_m) $$
in $\Pi^{(\sigma,S)}$ is satisfied in $(\sigma,S)$,
let us assume that its body is satisfied in $(\sigma,S)$, that is,
$(\sigma,S), a_1\ldots a_s \models  t_1, \ldots ,t_m$,
i.e.,
$(\sigma,S), a_1\ldots a_s \models  t_i$, for all $i=1, \ldots, m$.
By the same pattern of reasoning as above, we can show that
$h'_i \in R$, for all $i=1, \ldots, m$.
As rule (\ref{trad_causal_law}) is in $tr(\Pi)^R$, its body
is true in $R$, and $R$ is closed under $tr(\Pi)^R$, then
the head of  (\ref{trad_causal_law}), $(\neg) holds(f,s+1)$,
belongs to $R$.
Hence, by construction of $(\sigma, S)$, $ [a_1;\ldots; a_s]\bigcirc (\neg) f\in S$,
that is $(\sigma,S), a_1\ldots a_s \models H$,
namely, the head of the rule
$$[a_1,\ldots, a_s] ( H \leftarrow t_1, \ldots ,t_m) $$
is satisfied in $(\sigma,S)$.

\medskip

(ii) $S$ is minimal (in the sense of set inclusion)
among the $S'$ such that $(\sigma,S')$ is a partial interpretation
satisfying the rules in  $\Pi^{(\sigma,S)}$.

Assume by contradiction that $S$ is not minimal.
Then, there is a partial interpretation $(\sigma, S')$,
with $S'\subset S$, satisfying the rules in  $\Pi^{(\sigma,S)}$.

We show that we can construct an $R'\subset R$ such that $R'$ is closed under  $tr(\Pi)^R$.
We define $R'$ as $R$, but for the predicate $holds$, for which we have:
\begin{center}
$(\neg) holds(f,h) \in R'$ if and only if  $[a_1;\ldots; a_h](\neg)f \in S'$
\end{center}
It can be shown that $R'$ is closed under $tr(\Pi)^R$.
In fact, for each rule $r$ in  $tr(\Pi)^R$ whose body is true in $R'$, there is a rule $r'$ in $\Pi^{(\sigma,S)}$,
whose body is satisfied in $(\sigma,S')$.
As $(\sigma,S')$ satisfies all the rules in $\Pi^{(\sigma,S)}$, the head of $r'$ must be satisfied in $(\sigma,S')$.
By construction of $R'$, the head of $r$ belongs to $R'$.

As $R' \subset R$ and  $R'$ is closed under $tr(\Pi)^R$, $R$ is not minimal among the consistent sets of literals
closed under $tr(\Pi)^R$. This contradicts the hypothesis that $R$ is an answer set of $tr(\Pi)$.

\medskip
To prove that $R$ and $(\sigma, S)$ correspond to the same temporal model we can use the
same argument as for item (1).
\end{proof}

$\vspace{0.5mm}$

{\em Theorem \ref{SAT}}

Let $\Pi$ be the set of laws of a well-defined domain description,
$R$ an answer set of $tr(\Pi)$ and $\alpha$ a DLTL formula.
The temporal model $M_R=(\sigma,V)$ associated with $R$ satisfies $\alpha$ if and only if there is an answer set $R'$ of $tr'(\Pi)$
such that $R \subset R'$ and $sat(t\_alpha,0)\in R'$
(where  $t\_alpha$ is the term representing the formula $\alpha$
and $trans$ and $final$ encode the automata indexing the until formulas in $\alpha$).
\vspace{-0.3cm}
\begin{proof}

We first prove the "only if" direction  of the theorem.
We know by Theorem 1 that each answer set $R$
of $tr(\Pi)$ corresponds to a temporal answer set of $\Pi$
and, for each state $i=1,\ldots,k$, either
${holds(p,i)} \in R$ or ${\neg holds(p,i)} \in R$.
Let us consider the temporal model $M_R=(\sigma_R,V_R)$ associated with $R$, as defined in section 7.

We extend $R$ to define an answer set $R'$ of  $tr'(\Pi)$ as follows:
\begin{itemize}
\item
all the literals in $R$ belong to $R'$;

\item
for all subformulas $\beta$ of $\alpha$, for all states $h \in \{0,\ldots,k\}$:
\begin{equation} \label{sat_in_R}
sat(t\_beta, h)\in R' \mbox{ if and only if  } M_R, \tau_h\models \beta
\end{equation}
where $\tau_h= a_1\ldots a_h$ and $t\_beta$ is the term encoding the formula $\beta$.

\item
For each automaton ${aut}=(Q,\delta,Q_F)$ indexing an {\em until} formula in $\alpha$:
\begin{equation} \label{sat_in_R}
final(aut,q)\in R' \mbox{ if and only if  } q \in Q_f
\end{equation}
\begin{equation} \label{sat_in_R}
trans(aut,q_1,a,q_2)\in R' \mbox{ if and only if  } q_2 \in \delta(q_1,a)
\end{equation}

\end{itemize}

We can show that $R'$ is an answer set of $tr'(\Pi)$, i.e.,  (i) $R'$ is closed under $tr'(\Pi)^{R'}$
(ii)$R'$ is minimal among the consistent sets of literals closed under  $tr'(\Pi)^{R'}$.

(i) holds trivially for all the rules in $tr(\Pi)$. It has to be proved for all the rules defining the predicate $sat$.
We can procede by cases:

Let us consider the rule for fluents. Suppose $R'$ satisfies the body of a ground instance of the rule:

 $sat(F,S):- fluent(F), holds(F,S).$

\noindent
that is, $fluent(p) \in R'$ and $holds(p,h)\in R'$, for some fluent name $p$ and some $h \in \{1, \ldots, k\}$.
Then, $holds(p,h)\in R$, and thus $M_r, \tau_h \models p$.
By construction of $R'$, it must be: $sat(p,h) \in R'$.

Let us consider the first rule for until. Suppose $R'$ satisfies the body of a ground instance of the rule:

$sat(until(Aut,Q,Alpha,Beta),S):- final(Aut,Q),sat(Beta,S).$

\noindent
that is, for some $aut$ encoding a finite automaton ${\cal A}=(Q,\delta,Q_F)$, for some $q\in Q$,
for some formula $t\_beta$ and state $h$,
$final(aut,q) \in R'$ (i.e., $q \in Q_F$) and $sat(t\_beta,h)\in R'$.
By construction of $R'$, $M_R, \tau_h \models \beta$.
As $q$ is a final state of  the finite automaton ${\cal A}$, it must be that
$M_R, \tau_h \models \alpha {\cal U}^{{\cal A}(q)}\beta$.
Hence, by construction,

$sat(until(aut,q, t\_alpha, t\_beta),h)\in R'$.

\noindent
The other cases are similar.

(ii) We prove that $R'$ is minimal among the consistent sets of literals closed under  $tr'(\Pi)^{R'}$.
Let us suppose that $R'$ is  not minimal  and that there is an $R''\subset R'$ such that $R''$ is closed
with respect to $tr'(\Pi)^{R'}$.
There must be a literal $l \in R'-R''$.
$l$ cannot be a literal in $R$, as $R$ is an answer set of $tr(\Pi)$, and the definition of the predicates
in $R$ does not depend on the predicates $sat$, $trans$ and $final$ introduced in $tr'(\Pi)$.
Also, $l$ cannot be a $trans$ and $final$ literal, as these predicates are only defined
by ground atomic formulas, which must be all in $R''$.
Suppose there is $sat(t\_alpha,h) \in R'$ such that $sat(t\_alpha,h) \not \in R''$.

Using the fact that $R''$ is closed
with respect to $tr'(\Pi)^{R'}$,
it can be proved that,  for all the subformulas $\beta$ of $\alpha$,
if $M_R, \tau_h \models \beta$ then $sat(t\_beta,h) \in R''$.
The proof is by induction on the structure of $\beta$.

As $sat(t\_alpha,h) \in R'$, by construction of $R'$ it must be that
$M_R, \tau_h \models \alpha$.
Then, by the previous property, $sat(t\_alpha,h) \in R''$.
This contradicts the fact that $sat(t\_alpha,h) \not \in R''$.
Hence, $R'$ is an answer set of $tr(\Pi)$.

To conclude the proof of the ``only if'' part, it is easy to see that, from (\ref{sat_in_R}),
if $M_R, \varepsilon \models \alpha$  then
$sat(t\_alpha, 0)\in R'$, where $\varepsilon$ represents the empty sequence of actions.

\medskip

We have shown that, given an answer set $R$ of $tr(\Pi)$ satisfying $\alpha$
we can construct an answer set $R'$  of $tr'(\Pi)$ such that $sat(t\_alpha, 0)\in R'$.
To prove the "if" direction of the theorem,
let us assume that there is an answer set $R''$
of $tr'(\Pi)$
such that $R''$ extends $R$ and $sat(t\_alpha, 0)\in R''$.
We can show that $R''$ must coincide with $R'$ built above.
In fact, it can be easily proved that, for all subformulas $\beta$ of $\alpha$,

\begin{center}
 $sat(t\_beta, h)\in R''$  iff  $sat(t\_beta, h)\in R'$
\end{center}
The proof can be done by induction on the structure of $\beta$ (observe that both $R'$ and $R''$
extend $R$, which provides the evaluation of fluent formulas to be used by the sat predicate).
As $R''$ coincides with $R'$, if $sat(t\_alpha, 0)\in R''$ then
by  (\ref{sat_in_R}),
$M_R, \varepsilon \models \alpha$.
\end{proof}

\newpage

\begin{center}
\noindent {\Large \bf Appendix B}
\end{center}

\medskip

In this appendix we provide the encoding of BMC and Example \ref{es_mailbox} in DLV-Complex (https://www.mat.unical.it/dlv-complex).

\begin{verbatim}
state(0..#maxint).
laststate(N):- state(N), #maxint=N+1.

% general rules

occurs(A,S):- not ~occurs(A,S), action(A),state(S),laststate(L),S<=L.
~occurs(B,S):- occurs(A,S), action(A),state(S),action(B),A!=B.

next(S,SN):- state(S), laststate(LS), S<LS, SN=S+1.
-next(LS,S):- laststate(LS), next(LS,SS), state(S), state(SS), S!=SS.
next(LS,S):- laststate(LS), state(S), S<=LS, not -next(LS,S).
:- laststate(LS), next(LS,S), not eq_last(S).
	
diff_last(S):- state(S), S<#maxint, fluent(F),
               holds(F,S), -holds(F,#maxint).
diff_last(S):- state(S), S<#maxint, fluent(F),
               holds(F,#maxint), -holds(F,S).
eq_last(S):- state(S),  S<#maxint, not diff_last(S).

% The action theory makes use of the predicates:
%	action(A), fluent(FL), holds(FL,State)

% evaluation of DLTL formulas
% makes use of predicate formula(F)

% true
sat(true,S):- state(S).

% fluents
sat(F,S):- fluent(F), state(S), holds(F,S).

% not
sat(neg(Alpha),S):- formula(neg(Alpha)), state(S), not sat(Alpha,S).

% or
sat(or(Alpha1,Alpha2),S):- formula(or(Alpha1,Alpha2)), state(S),
                           sat(Alpha1,S).
sat(or(Alpha1,Alpha2),S):- formula(or(Alpha1,Alpha2)), state(S),
                           sat(Alpha2,S).

% until	
% An automaton is specified by the predicates
%		trans(Automaton,Q1,Action,Q2)  and
%		final(Automaton,Q)

sat(until(Aut,Q,Alpha,Beta),S):-
        formula(until(Aut,Q,Alpha,Beta)),
        final(Aut,Q),
        sat(Beta,S),
        state(S).
sat(until(Aut,Q,Alpha,Beta),S):-
        formula(until(Aut,Q,Alpha,Beta)),
        sat(Alpha,S),
        trans(Aut,Q,Act,Q1),
        action(Act),
        occurs(Act,S),
        next(S,S1),
        sat(until(Aut,Q1,Alpha,Beta),S1).
	

% derived operators and modalities	
%	ev(Alpha) means <>Alpha
%	diamond(Az,Alpha) means <Az>Alpha
%	box(Az,Alpha)  means  [Az]Alpha

sat(and(Alpha1,Alpha2),S):- formula(and(Alpha1,Alpha2)),
        state(S),
        sat(Alpha1,S), sat(Alpha2,S).
	
sat(impl(Alpha1,Alpha2),S):- formula(impl(Alpha1,Alpha2)),
        state(S),
        not sat(Alpha1,S).
sat(impl(Alpha1,Alpha2),S):- formula(impl(Alpha1,Alpha2)),
        state(S),
        sat(Alpha2,S).
	
sat(diamond(A,Alpha),S):- formula(diamond(A,Alpha)),
        action(A), state(S),
        occurs(A,S),
        next(S,SN),
        sat(Alpha,SN).
	
sat(ev(Alpha),S):- formula(ev(Alpha)),
        state(S),
        sat(Alpha,S).
sat(ev(Alpha),S):- formula(ev(Alpha)),
        state(S),
        next(S,SN),
        sat(ev(Alpha),SN).
	
sat(box(A,Alpha),S):- formula(box(A,Alpha)),
        action(A), state(S), action(B), formula(Alpha),
        occurs(B,S),
        A!=B.
sat(box(A,Alpha),S):- formula(box(A,Alpha)),
        state(S),
        occurs(A,S),
        next(S,SN),
        sat(Alpha,SN).

% the following rules define all subformulas of a given formula

formula(F):- formula(neg(F)).
formula(F1):- formula(or(F1,F2)).
formula(F2):- formula(or(F1,F2)).
formula(F1):- formula(until(Aut,Q,F1,F2)).
formula(F2):- formula(until(Aut,Q,F1,F2)).
formula(until(Aut,Q1,Alpha,Beta)):- formula(until(Aut,Q,Alpha,Beta)),
         trans(Aut,Q,Act,Q1).
formula(F1):- formula(and(F1,F2)).
formula(F2):- formula(and(F1,F2)).
formula(F1):- formula(impl(F1,F2)).
formula(F2):- formula(impl(F1,F2)).
formula(F):- formula(diamond(A,F)).
formula(F):- formula(ev(F)).
formula(F):- formula(box(A,F)).

% Encoding of Example 2

room(a).
room(b).

action(begin).
action(sense_mail(R)):- room(R).
action(deliver(R)):- room(R).
action(wait).

fluent(mail(R)):- room(R).

% action effects

holds(mail(R),SN):-
        room(R), occurs(sense_mail(R),S), SN=S+1,
        not -holds(mail(R),SN).	
-holds(mail(R),SN):-
        room(R), occurs(deliver(R),S), SN=S+1.

% persistency

holds(F,SN):-
        holds(F,S),
        SN=S+1,
        not -holds(F,SN).
-holds(F,SN):-
        ~holds(F,S),
        SN=S+1,
        not holds(F,SN).
	
%preconditions

:- occurs(deliver(R),S), -holds(mail(R),S).
:- occurs(wait,S), holds(mail(R),S).

%initial state

holds(mail(R),0):- room(R), not -holds(mail(R),0).
-holds(mail(R),0):- room(R), not holds(mail(R),0).

% temporal constraints

formula(diamond(begin,true)).

:- not sat(diamond(begin,true),0).

formula(neg(ev(neg(box(begin,until(aut,q1,true,true)))))).

trans(aut,q1,sense_mail(a),q2).
trans(aut,q2,sense_mail(b),q3).
trans(aut,q3,deliver(a),q4).
trans(aut,q3,deliver(b),q4).
trans(aut,q3,wait,q4).
trans(aut,q4,begin,q5).
final(aut,q5).

:- not sat(neg(ev(neg(box(begin,until(aut,q1,true,true))))),0).

% counterexample (negated property)

formula(ev(neg(impl(mail(b),ev(neg(mail(b))))))).

:- not sat(ev(neg(impl(mail(b),ev(neg(mail(b)))))),0).

\end{verbatim}

\newpage

\begin{center}
\noindent {\Large \bf Appendix C}
\end{center}

\medskip

In this appendix we report tests of our approach for bounded model checking of DLTL formulas,
in the line of the LTL BMC experiments in section 4 of \cite{Niemela03}.

In particular, we consider the dining philosophers problems and the LTL formulas in section 4 of \cite{Niemela03};
the relevant results are in Table 2 of that paper, columns $Int~n$ and $Int~s$, which provide, respectively,
the smallest integer such that a counterexample of length $n$ can be found using the interleaving semantics, and the time in seconds to find the counterexample.
The interleaving semantics is the relevant one since in
this paper we do not consider concurrent actions.

The general approach of the present paper can be directly mapped to the
DLV-Complex extension of the DLV system, as shown in Appendix B.
However, for a fairer comparison with the results in \cite{Niemela03},
we have tested a representation of
the dining philosophers problem, and of the LTL formulas to be verified,
in the DLV system rather than in its DLV-Complex extension.
Apart from not using parametric fluents and actions,
this means that, rather than using clauses (in section \ref{sec:translation}) such as

$sat(or(Alpha, Beta),S):- sat(Alpha,S).$

$sat(or(Alpha, Beta),S):- sat(Beta,S).$

\noindent
we provide, given the formula to be verified,
a unique name for the
formula and all its subformulas; and if a formula named $gamma$ is
the disjunction $alpha \vee beta$ of formulas named $alpha$ and $beta$,
we provide the clauses:

$sat(gamma,S):- sat(alpha,S).$

$sat(gamma,S):- sat(beta,S).$

\noindent
and similarly for other operators.
Such clauses can be easily generated automatically from the formula
to be verified.

Moreover, we have applied some
minor variation
of the general approach in section \ref{sec:translation} of our paper,
such as using DLV built-in predicates.

Table \ref{tab:philo} reports the results obtained on a Dell PowerEdge server with 2 Intel Xeon E5520 processors (2.26Ghz, 8M Cache) and 32 Gb of memory.

Column $n$ is the same as the $Int~n$ column in Table 2 of
\cite{Niemela03}, i.e.,
the smallest integer such that a counterexample of length $n$ can be found.
Column ``boundsmodels'' is the analogous of the $Int~s$ column in their
paper (except that we include the result for 12 philosophers); it provides the running times in seconds to find a counterexample,
running on our machine the code from http://www.tcs.hut.fi/kepa/experiments/boundsmodels/.
The last column provides the running times in seconds to find a counterexample running in DLV the programs enclosed.
The scalability of the approaches for such problems is similar,
and this provides some evidence that the approaches have similar practical relevance for
problems that can be represented easily in both of them.

\begin{table}[ht]
\centering

\begin{tabular}{| c | c | c | c |}
\hline
Problem & $n$ & boundsmodels & TemporalASP-DLV \\
\hline
DP(6) & 8  & 0.1 & 0.1\\
DP(8) & 10 & 1.4 & 2.4  \\
DP(10)& 12 & 29.1 & 115.7 \\
DP(12) & 14 & 7837.1 & 13036.2  \\
\hline
\end{tabular}
\caption{Dining philosophers results}
\label{tab:philo}
\end{table}

\end{appendix}


\begin{thebibliography}{}

\bibitem[\protect\citeauthoryear{Bacchus and Kabanza}{Bacchus and
  Kabanza}{1998}]{Bacchus98}
{\sc Bacchus, F.} {\sc and} {\sc Kabanza, F.} 1998.
\newblock Planning for temporally extended goals.
\newblock {\em Annals of Mathematics and Artificial Intelligence\/}~{\em 22},
  5--27.

\bibitem[\protect\citeauthoryear{Bacchus and Kabanza}{Bacchus and
  Kabanza}{2000}]{BacchusAIJ00}
{\sc Bacchus, F.} {\sc and} {\sc Kabanza, F.} 2000.
\newblock Using temporal logics to express search control knowledge for
  planning.
\newblock {\em Artificial Intelligence\/}~{\em 116,\/}~1-2, 123--191.

\bibitem[\protect\citeauthoryear{Balduccini and Gelfond}{Balduccini and
  Gelfond}{2003}]{Balduccini03}
{\sc Balduccini, M.} {\sc and} {\sc Gelfond, M.} 2003.
\newblock Diagnostic reasoning with {A}-prolog.
\newblock {\em Theory and Practice of Logic Programming\/}~{\em 3,\/}~4-5,
  425--461.

\bibitem[\protect\citeauthoryear{Baral and Gelfond}{Baral and
  Gelfond}{2000}]{Baral2000}
{\sc Baral, C.} {\sc and} {\sc Gelfond, M.} 2000.
\newblock Reasoning agents in dynamic domains.
\newblock In {\em Logic-Based Artificial Intelligence}. 257--279.

\bibitem[\protect\citeauthoryear{Baral and Zhao}{Baral and
  Zhao}{2007}]{Baral07}
{\sc Baral, C.} {\sc and} {\sc Zhao, J.} 2007.
\newblock Non-monotonic temporal logics for goal specification.
\newblock In {\em Proc. IJCAI 2007}. 236--242.

\bibitem[\protect\citeauthoryear{Bienvenu, Fritz, and McIlraith}{Bienvenu
  et~al\mbox{.}}{2006}]{Bienvenu06}
{\sc Bienvenu, M.}, {\sc Fritz, C.}, {\sc and} {\sc McIlraith, S.} 2006.
\newblock Planning with qualitative temporal preferences.
\newblock In {\em Proc. KR 2006}. 134--144.

\bibitem[\protect\citeauthoryear{Biere, Cimatti, Clarke, Strichman, and
  Zhu}{Biere et~al\mbox{.}}{2003}]{Cimatti03}
{\sc Biere, A.}, {\sc Cimatti, A.}, {\sc Clarke, E.~M.}, {\sc Strichman, O.},
  {\sc and} {\sc Zhu, Y.} 2003.
\newblock Bounded model checking.
\newblock {\em Advances in Computers\/}~{\em 58}, 118--149.

\bibitem[\protect\citeauthoryear{Bonatti}{Bonatti}{2004}]{Bonatti04}
{\sc Bonatti, P.} 2004.
\newblock Reasoning with infinite stable models.
\newblock {\em Artificial Intelligence\/}~{\em 156,\/}~1, 75--111.

\bibitem[\protect\citeauthoryear{Cla{\ss}en and Lakemeyer}{Cla{\ss}en and
  Lakemeyer}{2008}]{Classen:08}
{\sc Cla{\ss}en, J.} {\sc and} {\sc Lakemeyer, G.} 2008.
\newblock A logic for non-terminating {G}olog programs.
\newblock In {\em Proc. KR 2008}. 589--599.

\bibitem[\protect\citeauthoryear{{Dal Lago}, Pistore, and Traverso}{{Dal Lago}
  et~al\mbox{.}}{2002}]{Traverso02}
{\sc {Dal Lago}, U.}, {\sc Pistore, M.}, {\sc and} {\sc Traverso, P.} 2002.
\newblock Planning with a language for extended goals.
\newblock In {\em Proc. AAAI 2002}. 447--454.

\bibitem[\protect\citeauthoryear{D'Aprile, Giordano, Gliozzi, Martelli,
  Pozzato, and {Theseider Dupr{\'e}}}{D'Aprile et~al\mbox{.}}{2010}]{CLIMA2010}
{\sc D'Aprile, D.}, {\sc Giordano, L.}, {\sc Gliozzi, V.}, {\sc Martelli, A.},
  {\sc Pozzato, G.}, {\sc and} {\sc {Theseider Dupr{\'e}}, D.} 2010.
\newblock Verifying business process compliance by reasoning about actions.
\newblock In {\em CLIMA 2010, LNCS 6245}. 99--116.

\bibitem[\protect\citeauthoryear{Denecker, {Theseider Dupr\'{e}}, and {Van
  Belleghem}}{Denecker et~al\mbox{.}}{1998}]{Denecker98}
{\sc Denecker, M.}, {\sc {Theseider Dupr\'{e}}, D.}, {\sc and} {\sc {Van
  Belleghem}, K.} 1998.
\newblock An inductive definitions approach to ramifications.
\newblock {\em Electronic Transactions on Artificial Intelligence\/}~{\em 2},
  25--97.

\bibitem[\protect\citeauthoryear{Eiter, Faber, Leone, Pfeifer, and
  Polleres}{Eiter et~al\mbox{.}}{2003}]{Leone03}
{\sc Eiter, T.}, {\sc Faber, W.}, {\sc Leone, N.}, {\sc Pfeifer, G.}, {\sc and}
  {\sc Polleres, A.} 2003.
\newblock A logic programming approach to knowledge-state planning, {II}: The
  {DLV}$^{\mbox{{k}}}$ system.
\newblock {\em Artificial Intelligence\/}~{\em 144,\/}~1-2, 157--211.

\bibitem[\protect\citeauthoryear{Eiter, Faber, Leone, Pfeifer, and
  Polleres}{Eiter et~al\mbox{.}}{2004}]{Leone04}
{\sc Eiter, T.}, {\sc Faber, W.}, {\sc Leone, N.}, {\sc Pfeifer, G.}, {\sc and}
  {\sc Polleres, A.} 2004.
\newblock A logic programming approach to knowledge-state planning: Semantics
  and complexity.
\newblock {\em ACM Transactions on Computational Logic\/}~{\em 5,\/}~2,
  206--263.

\bibitem[\protect\citeauthoryear{Gelfond}{Gelfond}{2007}]{Gelfond}
{\sc Gelfond, M.} 2007.
\newblock {\em Handbook of Knowledge Representation, chapter 7, Answer Sets}.
\newblock Elsevier.

\bibitem[\protect\citeauthoryear{Gelfond and Lifschitz}{Gelfond and
  Lifschitz}{1988}]{Gelfond&Lifschitz:98}
{\sc Gelfond, M.} {\sc and} {\sc Lifschitz, V.} 1988.
\newblock The stable model semantics for logic programming.
\newblock In {\em Logic Programming, Proc. of the 5th Int. Conf. and
  Symposium}. 1070--1080.

\bibitem[\protect\citeauthoryear{Gelfond and Lifschitz}{Gelfond and
  Lifschitz}{1993}]{Gelfond&Lifschitz:93}
{\sc Gelfond, M.} {\sc and} {\sc Lifschitz, V.} 1993.
\newblock Representing action and change by logic programs.
\newblock {\em Journal of Logic Programming\/}~{\em 17}, 301--322.

\bibitem[\protect\citeauthoryear{Gerevini and Long}{Gerevini and
  Long}{2005}]{Gerevini05}
{\sc Gerevini, A.} {\sc and} {\sc Long, D.} 2005.
\newblock Plan constraints and preferences in {PDDL3}.
\newblock {\em Technical Report, Department of Electronics and Automation,
  University of Brescia, Italy\/}.

\bibitem[\protect\citeauthoryear{Giordano and Martelli}{Giordano and
  Martelli}{2006}]{AMAI06}
{\sc Giordano, L.} {\sc and} {\sc Martelli, A.} 2006.
\newblock Tableau-based automata construction for dynamic linear time temporal
  logic.
\newblock {\em Annals of Mathematics and Artificial Intelligence\/}~{\em
  46,\/}~3, 289--315.

\bibitem[\protect\citeauthoryear{Giordano, Martelli, and Schwind}{Giordano
  et~al\mbox{.}}{2001}]{GMS00}
{\sc Giordano, L.}, {\sc Martelli, A.}, {\sc and} {\sc Schwind, C.} 2001.
\newblock Reasoning about actions in dynamic linear time temporal logic.
\newblock {\em The Logic Journal of the IGPL\/}~{\em 9,\/}~2, 289--303.

\bibitem[\protect\citeauthoryear{Giordano, Martelli, and Schwind}{Giordano
  et~al\mbox{.}}{2007}]{GMS-JAL06}
{\sc Giordano, L.}, {\sc Martelli, A.}, {\sc and} {\sc Schwind, C.} 2007.
\newblock Specifying and verifying interaction protocols in a temporal action
  logic.
\newblock {\em Journal of Applied Logic\/}~{\em 5}, 214--234.

\bibitem[\protect\citeauthoryear{Giunchiglia}{Giunchiglia}{2000}]{Giunchiglia:%
KR2000}
{\sc Giunchiglia, E.} 2000.
\newblock Planning as satisfiability with expressive action languages:
  Concurrency, constraints and nondeterminism.
\newblock In {\em Proc. KR 2000}. 657--666.

\bibitem[\protect\citeauthoryear{Giunchiglia, Lee, Lifschitz, McCain, , and
  Turner}{Giunchiglia et~al\mbox{.}}{2004}]{Giunchiglia&al:2004}
{\sc Giunchiglia, E.}, {\sc Lee, J.}, {\sc Lifschitz, V.}, {\sc McCain, N.},
  {\sc }, {\sc and} {\sc Turner, H.} 2004.
\newblock Nonmonotonic causal theories.
\newblock {\em Artificial Intelligence\/}~{\em 153,\/}~1-2, 49--104.

\bibitem[\protect\citeauthoryear{Giunchiglia and Lifschitz}{Giunchiglia and
  Lifschitz}{1998}]{Giunchiglia&al:98}
{\sc Giunchiglia, E.} {\sc and} {\sc Lifschitz, V.} 1998.
\newblock An action language based on causal explanation: Preliminary report.
\newblock In {\em Proc. AAAI/IAAI 1998}. 623--630.

\bibitem[\protect\citeauthoryear{Giunchiglia and Traverso}{Giunchiglia and
  Traverso}{1999}]{Giunchiglia&Traverso99}
{\sc Giunchiglia, F.} {\sc and} {\sc Traverso, P.} 1999.
\newblock Planning as model checking.
\newblock In {\em Proc. 5th European Conf. on Planning (ECP'99)}. 1--20.

\bibitem[\protect\citeauthoryear{Heljanko and Niemel\"a}{Heljanko and
  Niemel\"a}{2003}]{Niemela03}
{\sc Heljanko, K.} {\sc and} {\sc Niemel\"a, I.} 2003.
\newblock Bounded {LTL} model checking with stable models.
\newblock {\em Theory and Practice of Logic Programming\/}~{\em 3,\/}~4-5,
  519--550.

\bibitem[\protect\citeauthoryear{Henriksen and Thiagarajan}{Henriksen and
  Thiagarajan}{1999}]{Henriksen99}
{\sc Henriksen, J.} {\sc and} {\sc Thiagarajan, P.} 1999.
\newblock Dynamic linear time temporal logic.
\newblock {\em Annals of Pure and Applied logic\/}~{\em 96,\/}~1-3, 187--207.

\bibitem[\protect\citeauthoryear{Kabanza, Barbeau, and St-Denis}{Kabanza
  et~al\mbox{.}}{1997}]{Kabanza}
{\sc Kabanza, F.}, {\sc Barbeau, M.}, {\sc and} {\sc St-Denis, R.} 1997.
\newblock Planning control rules for reactive agents.
\newblock {\em Artificial Intelligence\/}~{\em 95}, 67--113.

\bibitem[\protect\citeauthoryear{Leone, Pfeifer, Faber, Eiter, Gottlob, Perri,
  and Scarcello}{Leone et~al\mbox{.}}{2006}]{LeoneDLV}
{\sc Leone, N.}, {\sc Pfeifer, G.}, {\sc Faber, W.}, {\sc Eiter, T.}, {\sc
  Gottlob, G.}, {\sc Perri, S.}, {\sc and} {\sc Scarcello, F.} 2006.
\newblock The {DLV} system for knowledge representation and reasoning.
\newblock {\em ACM Transactions on Computational Logic\/}~{\em 7,\/}~3,
  499--562.

\bibitem[\protect\citeauthoryear{Levesque, Reiter, Lesp{\'e}rance, Lin, and
  Scherl}{Levesque et~al\mbox{.}}{1997}]{Levesque97}
{\sc Levesque, H.}, {\sc Reiter, R.}, {\sc Lesp{\'e}rance, Y.}, {\sc Lin, F.},
  {\sc and} {\sc Scherl, R.} 1997.
\newblock Golog: A logic programming language for dynamic domains.
\newblock {\em Journal of Logic Programming\/}~{\em 31,\/}~1-3, 59--83.

\bibitem[\protect\citeauthoryear{Lifschitz}{Lifschitz}{1990}]{Lifschitz:90}
{\sc Lifschitz, V.} 1990.
\newblock Frames in the space of situations.
\newblock {\em Artificial Intelligence\/}~{\em 46}, 365--376.

\bibitem[\protect\citeauthoryear{Panati and {Theseider Dupr\'{e}}}{Panati and
  {Theseider Dupr\'{e}}}{2000}]{Panati00}
{\sc Panati, A.} {\sc and} {\sc {Theseider Dupr\'{e}}, D.} 2000.
\newblock State-based vs simulation-based diagnosis of dynamic systems.
\newblock In {\em Proc. ECAI 2000}. 176--180.

\bibitem[\protect\citeauthoryear{Panati and {Theseider Dupr\'{e}}}{Panati and
  {Theseider Dupr\'{e}}}{2001}]{Panati01}
{\sc Panati, A.} {\sc and} {\sc {Theseider Dupr\'{e}}, D.} 2001.
\newblock Causal simulation and diagnosis of dynamic systems.
\newblock In {\em AI*IA 2001: Advances in Artificial Intelligence, LNCS 2175}.
  135--146.

\bibitem[\protect\citeauthoryear{Pistore and Traverso}{Pistore and
  Traverso}{2001}]{Pistore01}
{\sc Pistore, M.} {\sc and} {\sc Traverso, P.} 2001.
\newblock Planning as model checking for extended goals in non-deterministic
  domains.
\newblock In {\em Proc. IJCAI 2001}. 479--486.

\bibitem[\protect\citeauthoryear{Pistore, Traverso, and Bertoli}{Pistore
  et~al\mbox{.}}{2005}]{Pistore}
{\sc Pistore, M.}, {\sc Traverso, P.}, {\sc and} {\sc Bertoli, P.} 2005.
\newblock Automated composition of web services by planning in asynchronous
  domains.
\newblock In {\em Proc. ICAPS 2005}. 2--11.

\bibitem[\protect\citeauthoryear{Son, Baral, Tran, and McIlraith}{Son
  et~al\mbox{.}}{2006}]{Son06TCL}
{\sc Son, T.}, {\sc Baral, C.}, {\sc Tran, N.}, {\sc and} {\sc McIlraith, S.}
  2006.
\newblock Domain-dependent knowledge in answer set planning.
\newblock {\em ACM Transactions on Computational Logic\/}~{\em 7,\/}~4,
  613--657.

\bibitem[\protect\citeauthoryear{Son and Pontelli}{Son and
  Pontelli}{2006}]{Son06TPLP}
{\sc Son, T.~C.} {\sc and} {\sc Pontelli, E.} 2006.
\newblock Planning with preferences using logic programming.
\newblock {\em Theory and Practice of Logic Programming\/}~{\em 6,\/}~5,
  559--607.

\bibitem[\protect\citeauthoryear{Tu, Son, Gelfond, and Morales}{Tu
  et~al\mbox{.}}{2011}]{Morales08}
{\sc Tu, P.}, {\sc Son, T.}, {\sc Gelfond, M.}, {\sc and} {\sc Morales, R.}
  2011.
\newblock Approximation of action theories and its application to conformant
  planning.
\newblock {\em Artificial Intelligence\/}~{\em 175,\/}~1, 79--119.

\end{thebibliography}
\end{document}